\documentclass[journal]{vgtc}                     
\onlineid{1147} 
\usepackage{verbatim}

\usepackage{graphicx}
\usepackage{float}
\usepackage{algorithm}
\usepackage{algorithmicx}
\usepackage{algpseudocode}
\usepackage[caption=false]{subfig}
\usepackage{amsmath}
\usepackage{multirow}
\usepackage{bm}
\usepackage{amsfonts} 
\usepackage{color}  
\usepackage[dvipsnames]{xcolor}

\usepackage{booktabs} 
\usepackage{enumerate}
\usepackage{gensymb}
\usepackage{tabularx}
\usepackage{tabu,multirow,makecell,booktabs}

\usepackage{tabu}                      
\usepackage{booktabs}                  
\usepackage{lipsum}                    
\usepackage{mwe}                       
\usepackage{paralist,bbding,pifont}
\usepackage{amssymb}
\usepackage{amsmath}
\usepackage{enumitem}
\usepackage{multirow}

\usepackage{mathptmx}                  

\usepackage{marginnote}

\definecolor{red}{RGB}{193, 2, 19}
\definecolor{blue}{RGB}{50, 103, 185}

\newcommand{\pvis}[1]{\textcolor{black}{#1}}
\newcommand{\ques}[1]{\textcolor{red}{}}
\newcommand{\cl}[1]{\textcolor{red}{}}

\definecolor{red}{RGB}{193, 2, 19}
\definecolor{blue}{RGB}{50, 103, 185}

\usepackage{xcolor}
\usepackage{tcolorbox}

\usepackage{mathptmx}                  

\definecolor{titleblockcolor}{HTML}{353535}
\definecolor{textblockcolor}{HTML}{FFFFFF}
\newenvironment{block}[2][]{
  \begin{tcolorbox}[adjusted title=#2, fonttitle={\normalsize\bfseries}, colback={textblockcolor}, colframe={titleblockcolor}, coltitle={white}, arc=0pt,
  outer arc=0pt, left=1pt, right=1pt, fontupper=\normalsize, #1]
}{\end{tcolorbox}}
\newtcbox{\highlight}[1][red]
  {on line, arc = 0pt, outer arc = 0pt,
    colback = #1!10!white, colframe = #1!50!black,
    boxsep = 0pt, left = 1pt, right = 1pt, top = 2pt, bottom = 2pt,
    boxrule = 0pt, bottomrule = 1pt, toprule = 1pt}

\begin{document}

\title{VizDefender: Unmasking Visualization Tampering through Proactive Localization and Intent Inference}

\author{Sicheng~Song,
        Yanjie~Zhang,
        Zixin~Chen,
        Huamin~Qu,
        Changbo~Wang,
        and Chenhui~Li}

\authorfooter{
\vspace{-4px}
\item S. Song are both with East China Normal University and the Hong Kong University of Science and Technology E-mail:  csescsong@ust.hk.
\item Y. Zhang, Z. Chen, and H. Qu are with the Hong Kong University of Science and Technology E-mail: \{yzhangvj, zchendf\}@connect.ust.hk, huamin@ust.hk.
\item C. Wang, C. Li are with the School of Computer Science and Technology, East China Normal University, E-mail:  \{chli, cbwang\}@cs.ecnu.edu.cn.
\item This work was supported by the National Natural Science Foundation of China under Grants 62572191 and 62472178, and by the Natural Science Foundation of Shanghai Municipality under Grant 24ZR1418300.
\item Chenhui Li and Changbo Wang are the
corresponding authors.
\vspace{-8px}
}

\markboth{IEEE TRANSACTIONS ON VISUALIZATION AND COMPUTER GRAPHICS}%
{Song \MakeLowercase{\textit{et al.}}: VizDefender: Visualization Tampering Defense with Manipulation Intent Inference}

\abstract{
The integrity of data visualizations is increasingly threatened by image editing techniques that enable subtle yet deceptive tampering. Through a formative study, we define this challenge and categorize tampering techniques into two primary types: data manipulation and visual encoding manipulation. To address this, we present VizDefender, a framework for tampering detection and analysis. The framework integrates two core components: 1) a semi-fragile watermark module  that protects the visualization by embedding a location map to images, which allows for the precise localization of tampered regions while preserving visual quality, and 2) an intent analysis module that leverages Multimodal Large Language Models (MLLMs) to interpret manipulation, inferring the attacker's intent and misleading effects. Extensive evaluations and user studies demonstrate the effectiveness of our methods.
}


\teaser{
  \centering
    \includegraphics[width=0.97\linewidth ]{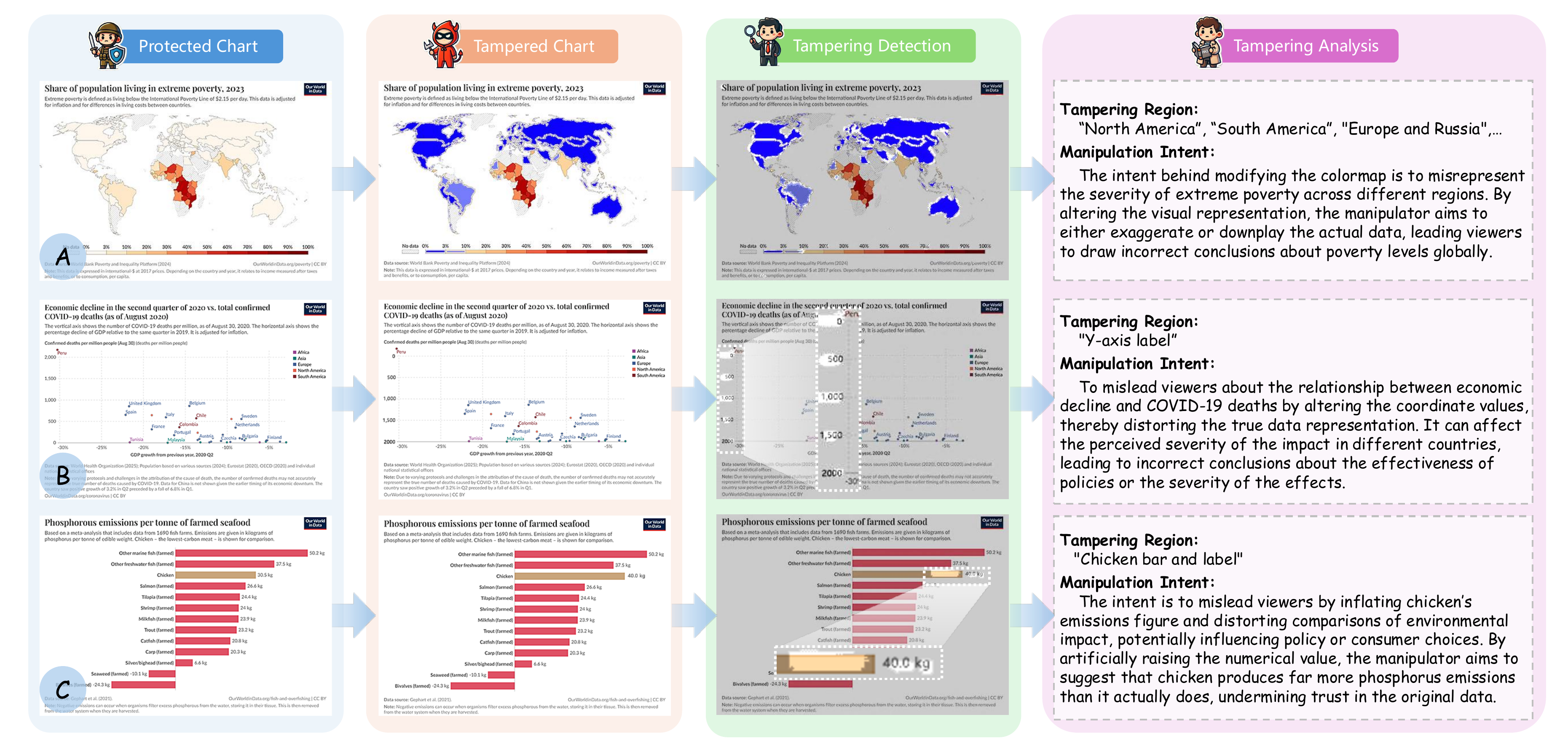}
          \vspace{-5px}
  \caption{\label{fig:teaser}This figure illustrates three different types of data tampering and their detection process. The Protected Visualization column shows the original visualizations with embedded location map: (A) Share of population living in extreme poverty in 2023~\cite{owid-poverty}; (B) Economic decline in the second quarter of 2020 vs. total confirmed COVID-19 deaths as of August 2020~\cite{owid-coronavirus}; (C) Phosphorous emissions per tonne of farmed seafood~\cite{owid-fish-and-overfishing} The Tampered Visualization column in the displays the tampered versions of these visualizations, which include modifying colormap, coordinate values, and data point values. The Tampering Detection column highlights the detected tampered areas using overlay masks. The Tampering Analysis column provides detailed explanations of each tampering region and its intent, revealing how the alterations mislead viewers and compromise the accuracy of the data.}
}

\maketitle




\graphicspath{{figs/}{figures/}{pictures/}{images/}{./}} 


\section{Introduction}
The integrity of data visualizations is crucial for informed decision-making. However, the digital age has brought an insidious threat: tampered visualizations. These are existing visualizations subtly manipulated—by altering original data or visual encodings—to distort the truth. This distinction is vital because tampered visualizations, by leveraging a authentic foundation, are more easily produced and disseminated than entirely fabricated visuals, and their ``half-truth'' nature makes them difficult to discern from genuine information.

This poses a significant challenge for social media platforms, whose content moderation teams must protect users from widespread misinformation. As the World Economic Forum's Global Risks Report 2025 highlights, misinformation and disinformation stand as the leading global concern in the near future, a top spot they have held for two years running~\cite{WEF_GlobalRisks2025}. Meanwhile, AI is also improving people's critical thinking. Some empirical studies~\cite{augenstein2024factuality, ortiz2025fact} have confirmed that more and more people will use chatbots like Grok to check facts on social platforms such as Twitter. However, most of these fact checking tools are aimed at textual information and natural images. The ease with which image editing software and generative AI tools now enable such convincing alterations to existing visualizations has lowered the barrier for malicious actors. For social media content moderation teams who need to rapidly filter massive volumes of content and social media users who are often the producers and consumers of these visuals,  manually assessing a single claim can take a professional several hours or days~\cite{quelle2024perils}. This urgent need for automated and precise detection for visual manipulation.

Watermarking is common on social media for copyright protection, but its application in visualization community has focused on data retrieval. Techniques like VisCode~\cite{zhang2020viscode}, InvVis~\cite{ye2023invvis}, and VisGuard~\cite{ye2025visguard} embed robust watermarks (e.g., QR codes) that are designed to survive modifications to allow users to trace back to the original data. However, their robustness makes them unsuitable for precisely localizing tampered regions, which requires a semi-fragile signal. They also require complex workflows to store raw data and perform manual comparison. Meanwhile, general-purpose tampering detection networks~\cite{wu2019mantra, dong2022mvss} and even proactive watermarking frameworks like EditGuard~\cite{zhang2024editguard} are tailored for natural images and fail to address the unique structural properties of visualizations or infer manipulation intent.

To address these gaps and empower social media content moderation teams, we propose \textbf{VizDefender}, a framework for proactive tampering detection and intent analysis. Fig.~\ref{fig:teaser} illustrates this entire workflow across three examples (A, B, C), walking through the full process from protection to analysis. Our framework consists of two main components.
First, a semi-fragile watermark module leverages the fragility and locality properties of invertible neural networks (INNs) to embed a signal into the original \textit{Protected Chart}.  Semi-fragile watermarks are designed to tolerate benign operations like compression, but break predictably upon malicious edits to localize the tampered area. When a \textit{Tampered Chart} is fed into our system, our method allows the module to precisely localize alterations (\textit{Tampering Detection}) while remaining robust to benign changes like file compression. Second, an intent analysis and interpretation module (\textit{Tampering Analysis}) uses a novel two-agent MLLM pipeline to semantically interpret these detected regions. This pipeline employs constrained, rule-based reasoning—mapping components like axes or bars to specific manipulation methods—to analyze the tamper, infer the manipulator's intent, and explain its misleading effects. This constrained approach reduces model hallucination and improves interpretive accuracy. The deployment of such a proactive framework can help build a corpus of real-world tampering cases, paving the way for future data-driven passive detection models.

Our contributions include three aspects:

\begin{compactitem}
    \item [(1)] We conduct a formative interview study to reveal the definition and challenges of tampered visualizations and define a new problem of visualization tampering detection and analysis.
    \item [(2)] We introduce a novel framework for detecting and analyzing tampered visualizations using semi-fragile watermarking combined with intelligent interpretation leveraging MLLMs.
    \item [(3)] We validate the effectiveness of our framework through extensive quantitative evaluations against baseline methods and a user study.
\end{compactitem}

\section{Related Work}
Our work is situated at the intersection of three research areas. We first review how the visualization community has traditionally addressed misinformation, then examine existing image tampering detection techniques from computer vision, and finally discuss information steganography as the foundation for our proactive defense approach.

\subsection{Misinformation Visualization}
Misinformation in visualizations can distort public perception. Seminal works like Huff's \textit{How to Lie with Statistics}~\cite{huff1993lie} and Tufte's critique of ``\textit{chart junk}''~\cite{tufte1983visual} established a foundational understanding of how visual design choices can mislead.

Building on this, modern research has created extensive taxonomies of misleading design practices, such as truncating axes~\cite{correll2020truncating, pandey2015deceptive} or creating deceptive visual encodings~\cite{lo2022misinformed, lan2024junk}. To combat these issues, various automated tools and frameworks~\cite{chen2021vizlinter,mcnutt2018linting,lei2023geolinter} have been developed to detect design flaws at the source code level.  With the rise of reverse visualization engineering~\cite{poco2017reverse, savva2011revision}, the pipelines of extracting raw data from bitmaps and pointing out design flaws or misleading areas has been implemented in many chart types such as bar charts~\cite{Schlieder2024metacognition}, line charts~\cite{fan2022annotating}, scatter plots~\cite{shi2024under}, node-link diagrams~\cite{9720180,song2022graphdecoder,10734252}, etc. In addition to these works, there are some studies focused on labeling potential misleaders~\cite{hopkins2020visualint} on graphs and strategies to explain common misleaders to the public~\cite{lo2023change}.

However, these works focus on flaws introduced during the initial chart creation. It largely overlooks the growing threat of post-creation tampering, where a well-designed visualization is maliciously altered at the image level. While recent studies have explored using MLLMs to identify misleading designs in chart images~\cite{10679256,10857634,chen2025unmasking}, detecting falsified data in a tampered static image remains a challenge. 
This specific gap motivates the need for dedicated tampering detection solutions.

\subsection{Image Tampering Detection}
While the visualization community has focused on design, the computer vision field has developed methods for image-level manipulation detection, though almost for natural images.

These methods can be broadly categorized as passive or proactive. Passive methods, such as ManTra-Net~\cite{wu2019mantra} and MVSS-Net~\cite{dong2022mvss}, search for tampering artifacts without any prior information, but their performance depends heavily on the types of images and manipulations seen during training. Proactive methods, like MaLP~\cite{asnani2023malp} and EditGuard~\cite{zhang2024editguard}, pre-emptively embed a digital watermark, to verify integrity of images.

Despite their sophistication, these techniques are not directly applicable to data visualizations. The subtle, semantic nature of chart tampering—where changing a few pixels can alter a data point's value—is fundamentally different from the structural anomalies typically found in manipulated natural images. Existing models lack the semantic understanding of chart components, highlighting the need for a domain-specific proactive defense. This leads us to explore the specific application of information steganography in visualization images.

\subsection{Information Steganography}

Information Steganography is a technique used to embed hidden information into various types of carriers such as text, audio, video, and images. Among these carriers, images are the most widely used. Traditional methods mainly focus on the spatial domain~\cite{mielikainen2006lsb,kawaguchi1999principles,pevny2010using} or the transformation domain~\cite{almohammad2008high,swanson1997multiresolution,zhu1999multiresolution}. With the development of deep learning, techniques such as autoencoders~\cite{zhu2018hidden,baluja2017hiding} and GANs (Generative Adversarial Networks)~\cite{goodfellow2014generative} have been employed to optimize the embedding process, achieving better performance. Recently, Flow-based models have further advanced this field~\cite{ma2022towards, fang2023flow, ye2024pprsteg}.

In visualizations, embedding metadata into visualization images has gained attention to maintain interactivity and integrity. For instance, Hota et al.\cite{hota2019embedding} embedded digital watermarks into scientific visualizations to protect metadata, and Yang et al.\cite{yang2019embedding} embedded watermarks in 3D models. For information visualization, , this technique has been adopted for data integrity and retrieval. For example, systems like VisCode~\cite{zhang2020viscode}, Chartem~\cite{fu2020chartem}, ChartStamp~\cite{fu2022chartstamp}, InvVis~\cite{ye2023invvis}, and VisGuard~\cite{ye2025visguard} embed information (e.g., QR codes linking to source data) into charts. The primary goal of these methods is data retrieval. Although they can be partially applied to tampering detection, they require additional costs to store and compare the original data, and trivial morphological comparison methods make it difficult for automated processes to accurately locate the tampering regions. Another gap is that these works do not use the semantic information of the visualizations to infer tampering.

Our work addresses this critical gap by developing an end-to-end framework for detecting and localizing tampering regions and infer manipulation intents in visualizations.

\section{Formative Interview Study}

We conducted a formative interview study to identify and categorize common tampering types performed on existing visualization images. The goal was to validate and refine the taxonomy of visual manipulation types~\cite{lan2024junk, chen2025unmasking}, ensuring that our research focuses on the most impactful and prevalent forms of manipulation.

\textbf{Participants:} Our study recruited $20$ volunteers  ($\mu_{age}=27.15$ years, $STD_{age}=6.54$, $10$ females and $10$ males), most of whom were postgraduate students majoring data visualization, computer vision, or related fields. The interviewees also included two professors with over 10 years of experience in data visualization. All participants possessed foundational visualization literacy.

\textbf{Procedure:} We conducted semi-structured individual interviews online, each lasting approximately $30$ minutes with additional time allowed if needed for participants to fully express their views. After obtaining informed consent, we presented the concepts of visualization tampering and misinformation. Participants were then asked to review an existing taxonomy~\cite{lan2024junk} and a benchmark~\cite{chen2025unmasking}, providing feedback on which manipulation types could be achieved through image editing and suggesting new ones. The taxonomy offers a broad categorization of real-world misleading designs from a public perspective, while the benchmark provides a set of standardized D3-synthesized misleading chart examples, allowing our interview to be grounded in both in-the-wild flaws and controlled, systematic cases. The interviews focused on identifying visualization tampering, categorizing tampering types, their impact, and the challenges in their detection.

\begin{figure}[!btp]
    \centering
    \includegraphics[width=\linewidth]{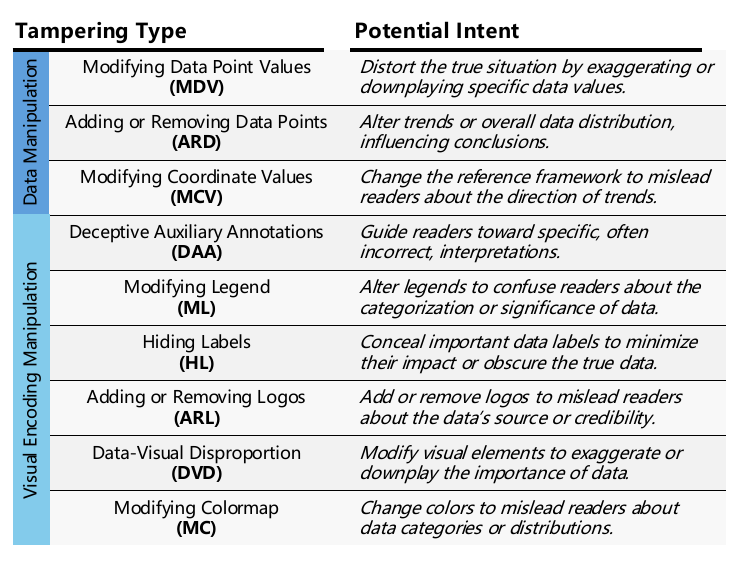}
    \vspace{-20px}
    \caption{We summarized 9 identified common tampering types into two main categories based on their potential image manipulation intents. The bold letters are the abbreviations of each type.}
    \label{fig:tamperType}
    \vspace{-15px}
\end{figure}

\textbf{Analysis:} We analyzed the interview data using thematic analysis~\cite{braun2006using}, which yielded the following insights.

\textbf{Tampered visualizations are often more deceptive than traditional misleading visualizations.} Participants noted a difference in the perceived intention behind tampered visualizations compared to traditional misleading visualizations. While misleading visualizations may result from poor design choices or unintentional biases, tampered visualizations are deliberately altered to distort data and manipulate viewers' perceptions. Multiple participants (e.g., P1, P2, P6) remarked that tampered visualizations are specifically designed to deceive on purpose. As P1 stated, ``\textit{Tampered visualizations are designed to deceive on purpose. It's not just about poor design; it's a deliberate act to mislead.}'' Participants also stated the relationship between tampered visualizations and misinformation visualizations. They observed that tampered visualizations are a significant subset of misinformation visualizations, including fake annotations, fake data, and biased interpretations. As P14 commented, ``\textit{Tampered visualizations are one part of the bigger picture of misinformation. They are insidious because they involve direct alterations to the visual elements, making it hard to trace the source of the manipulation.}'' This insight was supported by several other participants (P7, P11), indicating the need to consider tampered visualizations within the diverse types of tampering.

\textbf{Detecting tampered visualizations is more challenging.} Participants highlighted that tampered visualizations, which are altered post-creation using image editing software, can be significantly more challenging to identify due to the subtlety of image differences. P3 noted, ``\textit{It's much harder to spot tampered visualizations because the changes are often very subtle and seamlessly integrated into the original image. You need to have a keen eye and a good understanding of data visualization to catch them.}'' P8 echoed this sentiment, stating, ``\textit{Tampered charts can be very sophisticated. With advanced image editing software, it's possible to make changes that are nearly undetectable by human eyes.}'' Participants also pointed out that image editing tools have become increasingly powerful and accessible, allowing non-experts to make convincing tampering to visualizations. This insight points to the need for techniques to detect tampered visualizations effectively.

\begin{figure*}[tb]
\centering
\setlength{\fboxrule}{0.4pt}
\setlength{\fboxsep}{0cm}
\includegraphics[width=\linewidth]{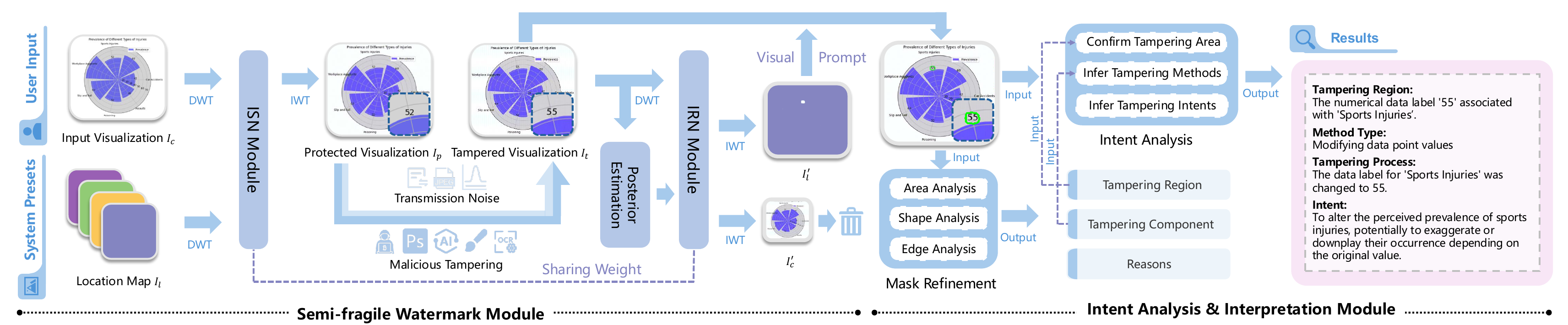}\ 

\vspace{-5px}
\caption{\label{fig:pipeline}
Overview of VizDefender pipeline. Our framework consists of two components: Semi-fragile Watermark Module, and Intent Analysis \& Interpretation. In the watermark processing stage, a flow-based model embeds the location map into the input visualization image, generating a protected visualization image that can be transmitted through networks. After potential tampering and transmission noise, an IRN module with posterior estimation extracts the watermark to generate the visual prompt on the tampered visualization. The intent analysis stage then processes both the tampered visualization with visual prompt through refinement and analysis modules, outputting intent results.}
    
\vspace{-5px}
    
\end{figure*}

\textbf{Understanding the manipulator's intent is crucial.} A crucial insight from the interviews was the importance of understanding the intent behind the manipulation. Participants emphasized that the intent of the manipulator is a key factor that distinguishes visualization tampering from other forms of image types. P19 explained, ``\textit{The intent behind tampered visualizations is more diverse than those in natural images.}'' This sentiment was echoed by P12, who stated, ``\textit{Understanding the intent behind the manipulation is crucial. It helps us identify the harmfulness of tampering and enhance the data literacy of the public.}'' This insight highlights the need to focus on the manipulator's intent when developing strategies to detect and prevent tampered visualizations. 

\textbf{Research scope of tampering types.} Through the interviews, several common tampering types were identified, which were categorized into two primary groups: data manipulation and visual encoding manipulation, which is as shown in Fig.~\ref{fig:tamperType}.  Data manipulation includes modifying data point values, adding or removing data points, and tampering coordinate values. For example, in bar charts, this involves modifying the values on bars to achieve wrong data insights. In line charts, this can modify the values on the coordinate axis to alter trends and data distribution. Modifying coordinate values can change the reference framework, leading readers to incorrect interpretations of data trends. These manipulations are intended to distort the actual situation by either exaggerating or downplaying specific data values, influencing conclusions drawn from the visualizations.

Visual encoding manipulation, on the other hand, involves deceptive auxiliary annotations, modifying legends, hiding labels, adding or removing logos, data-visual disproportion, and changing color maps. Deceptive auxiliary annotations can guide readers toward specific, often incorrect interpretations. In scatter plots, it can involve adding deceptive cluster circles to mislead perception of categories. Hiding labels can obscure important data, minimizing their impact. Adding or removing logos can mislead readers about the data's source or credibility. Data-visual disproportion involves altering visual elements. For instance, in bar charts, visual encoding manipulation might include changing the height of bars to exaggerate smaller values or altering the color map. Modifying color maps can mislead readers about data categories or distributions by changing colors or inverting color scales.

The refined taxonomy of visualization manipulation types will serve as a foundation for developing our intent inference model to address the challenges posed by tampered visualizations.

\section{Methods}
Our framework as shown in Fig.~\ref{fig:pipeline} consists of two main components: 

(1) \textit{Semi-fragile Watermark Module:} The watermark processing stage includes both embedding and extraction models. Unlike traditional robust watermarks~\cite{ye2023invvis, ye2025visguard} that aim to survive any transformation, our semi-fragile watermark is designed to be sensitive to image manipulations while remaining legitimate operations. The watermark is designed to maintain visual quality while ensuring detectability and spatial-related of various tampering operations. We also employed a posterior estimation~\cite{zhang2024editguard} leverages prompts to strengthen watermark extraction under various degradations.

(2) \textit{Intent Analysis and Interpretation:} The interpretation module leverages MLLMs to provide analysis of detected masks. This component processes both the tampering masks and visualization context to infer manipulation techniques and their potential misleading effects. The analysis helps users understand not just where tampering occurred, but also the intent behind the manipulation.

\begin{table}[btp]
\centering
\vspace{-10px}
\caption{\label{training_set}Distribution of different visualization types in our training dataset.}
\vspace{-10px}
    \includegraphics[width=\linewidth]{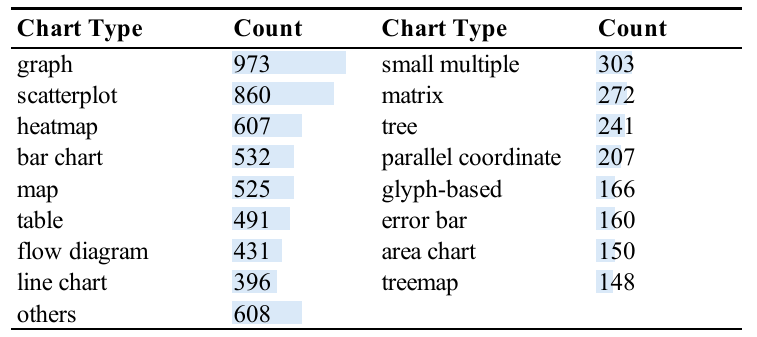}
\vspace{-20px}
\end{table}

\subsection{Training Dataset}
For training our semi-fragile watermark models, we curate a specialized dataset from VisImages~\cite{deng2024visimages}, a visualization corpus containing $35,096$ images from IEEE VIS publications. VisImages features expert-designed visualizations with diverse styles and complex layouts, making it suitable for training watermark-related models in visualization contexts. To ensure high-quality training data, we filter the original corpus based on image resolution, selecting only images with both height and width larger than $512$ pixels. This filtering process results in $7,070$ visualizations, which are then split into training ($5,656$ images, $80\%$) and validation ($1,414$ images, $20\%$) sets. The selected visualizations cover various types as shown in Table~\ref{training_set}, ensuring our model's capability to handle different visualization types. 

To accelerate convergence and enhance the robustness of our model, we initially pre-train it on the COCO dataset~\cite{lin2014microsoft}, which provides a diverse set of images and helps the model learn generalizable features. We used the Adam optimizer with an initial learning rate of $1 \times 10^{-4}$. For the pre-training phase, the model was trained for 250K iterations with a batch size of $4$. Subsequently, for the fine-tuning phase on our curated VisImages dataset, we trained for an additional 10K iterations with an initial learning rate of $1 \times 10^{-5}$ and a batch size of $4$.

\subsection{Semi-fragile Watermark Module}
Recent advances in image-to-image steganography using invertible neural networks (INNs)~\cite{lu2021large} have demonstrated impressive performance in information hiding. Through our observations, we discover two properties of INN-based steganography that make it suitable for tampering detection: (1) Fragility: when the container image undergoes tampering, the extracted watermark also exhibits damages, and (2) Locality: the damage to the extracted watermark precisely corresponds to the spatial location of the modifications, showing position-wise correlation.

Instead of trying to overcome these inherent characteristics as in previous works~\cite{ye2025visguard, ye2023invvis}, we leverage INN's natural fragility and locality for tampering detection. By designing a semi-fragile watermark that is sensitive to malicious manipulations while remaining stable under legitimate operations, we can  localize tampered regions through comparing the extracted watermark with the original embedded one. When users input a clean visualization image $I_c \in \mathbb{R}^{H \times W \times 3}$ into our system, along with a pre-defined location map $I_l \in \mathbb{R}^{H \times W \times 3}$ (typically a solid-color image), our framework processes them through an ISN (Image Steganography Network) module. The ISN first applies DWT (Discrete Wavelet Transform) to decompose both $I_c$ and $I_l$ into frequency domains. Through invertible blocks, the network embeds $I_l$ into $I_c$ to generate a protected visualization $I_p \in \mathbb{R}^{H \times W \times 3}$. During network transmission, $I_p$ may encounter various malicious manipulations and normal compressions, resulting in a tampered visualization $I_t$. The IRN (Image Revealing Network) module then processes $I_t$ through invertible blocks and posterior estimation to extract the watermark $I'_l$ and the reconstructed visualization $I'_c$. The damaged areas of $I'_l$ are the area where the visualization is detected to be potentially tampered with.

\subsubsection{Invertible Network}
The key characteristic of INNs is that they allow reconstruction of inputs from their outputs through an inverse transformation. The core components of our INN are invertible blocks with affine coupling layers. These blocks follow a design pattern that ensures invertibility while maintaining expressive power: they split the input features into two parts and transform them alternately, with each part's transformation depending only on the other part. This design allows both forward and inverse computations to be tractable and computationally efficient.

\begin{figure}[tbp]
    \centering
    \vspace{-5px}
\includegraphics[width=\linewidth]{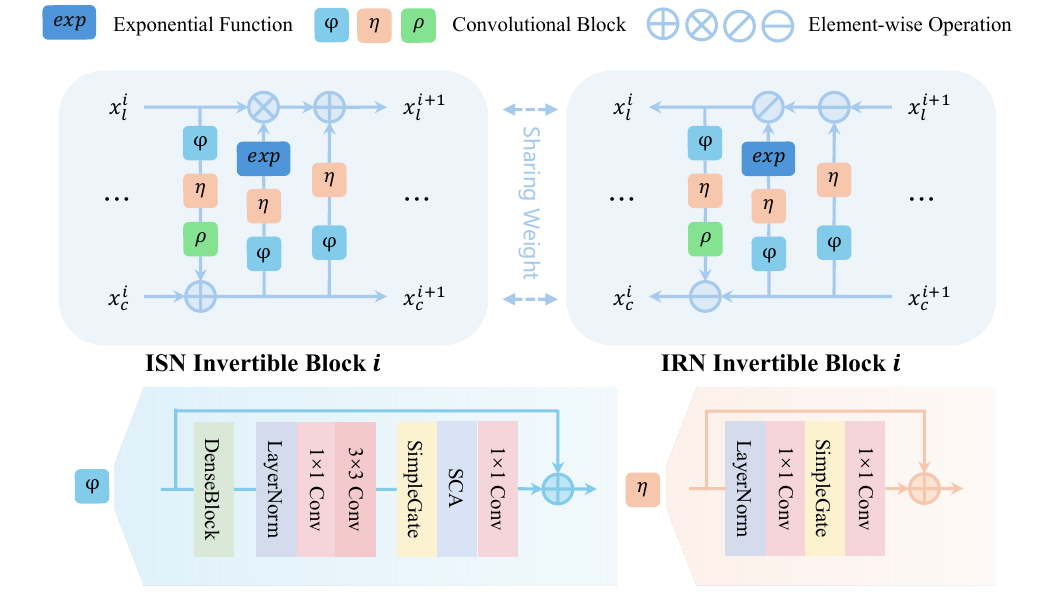}
    
\vspace{-5px}
    \caption{The architecture of invertible blocks in ISN Module and IRN Module. Key components such as $\phi$, $\eta$, $\exp$, and $\rho$ are integrated to enhance the transformation capabilities.}
    
    \label{fig:INN}
    
\vspace{-10px}
\end{figure}

To improve embedding quality and reduce visual artifacts, we first apply DWT to decompose both the input visualization $I_c$ and location map $I_l$ into frequency domains. The location map is a simple, predefined image (e.g., a checkerboard pattern) embedded as a fragile baseline. Any deviation in the extracted map directly reveals the spatial location of tampering. This frequency-domain decomposition helps our network concentrate watermark information in high-frequency components while preserving the visual structure in low-frequency components~\cite{guan2022deepmih}. The transformed features then pass through a series of invertible blocks with affine coupling layers. For the $i$-th invertible block as shown in Fig.~\ref{fig:INN}, the forward propagation of ISN is defined as:

\begin{equation}
\begin{aligned}
x_l^{i+1} &= x_l^i \odot \text{exp}(\eta(\phi(x_c^{i+1}))) + \eta(\phi(x_c^{i+1})) \\
x_c^{i+1} &= x_c^i + \rho(\eta(\phi(x_l^{i})))
\end{aligned}
\end{equation}
where $x_c^i$ and $x_l^i$ represent the features of the visualization image and location map in the $i$-th layer respectively, $\odot$ denotes element-wise multiplication, and $\text{exp}(\cdot)$ is the exponential function. The transformation functions $\phi(\cdot)$ and $\eta(\cdot)$ are implemented as a five-layer dense block~\cite{huang2017densely} augmented with a lightweight feature interaction module~\cite{chen2022simple}. 
In reverse, the revealing operation of IRN module can be computed as:

\begin{equation}
\begin{aligned}
x_l^i &= (x_l^{i+1} - \eta(\phi(x_c^{i+1})) \odot \text{exp}(\eta(\phi(x_c^{i+1})))  \\
x_c^i &= x_c^{i+1} - \rho(\eta(\phi(x_l^{i})))
\end{aligned}
\end{equation}

\subsubsection{Posterior Estimation}
\begin{figure}[tbp]
    \centering
    \includegraphics[width=\linewidth]{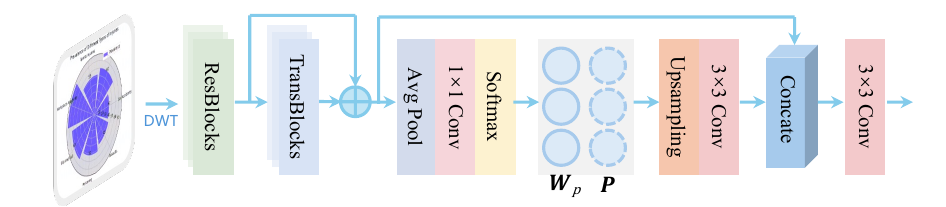}
    \vspace{-20px}
    \caption{The posterior estimation model.}
    \vspace{-10px}
    \label{fig:PEM}
\end{figure}

To keep the semi-fragile watermark robust to some normal image degradations, we add a posterior estimation model~\cite{zhang2024editguard} to IRN, which is shown to predict missing high-frequency information. The model (Fig.~\ref{fig:PEM}) helps our INNs to accurately recover the location map $I_l$ from the tampered visualization $I_t$. The model first extracts local and non-local features $F_t$ from the tampered visualization $I_t$ using a combination of residual blocks and transformer blocks. This can be formulated as:
\begin{equation}
      F_t = \zeta_2(\zeta_1(\text{DWT}(I_t))) + \zeta_1(\text{DWT}(I_t))
\end{equation}
where $\zeta_1(\cdot)$ denotes $8$ residual blocks~\cite{he2016deep} and $\zeta_2(\cdot)$ denotes $4$ transformer blocks~\cite{zamir2022restormer}. Pre-defined degradation prompts $P = [P_1, P_2, P_3]$ are learnable embedding tensors that represent three types of typical transmission noise including Gaussian Noise, JPEG, and Poisson Noise. These prompts are combined with the extracted features using dynamic weight coefficients $W_p$ obtained via a average pooling layer, a $1\times1$ convolution layer and a softmax operation. The fused features and degradation prompts generate the estimated posterior, which initializes the invertible blocks in the IRN, enabling accurate recovery of $I_l$.

\subsubsection{Loss Function}

Our training objective is twofold: ensuring the visual fidelity of the protected visualization ($I_p$) to the original ($I_c$), and guaranteeing the accurate extraction of the location map ($I'_l$) from an untampered image. To achieve this, we define a total loss $\mathcal{L}_{\text{total}}$ as a weighted sum of an encoding loss and an extraction loss. The encoding loss is the Mean Squared Error (MSE) between $I_p$ and $I_c$, ensuring visual similarity. The extraction loss is the L1 loss between the original location map $I_l$ and the extracted map $I'_l$, ensuring accurate recovery. The total loss function can be formulated as:
\begin{equation}
\label{equ:total_loss}
    \mathcal{L}_{\text{total}} = \alpha \mathcal{L}_{\text{enc}} + \beta \mathcal{L}_{\text{ext}} = \alpha \mathcal{L}_{\text{mse}}(I_p, I_c) + \beta \mathcal{L}_{\text{L1}}(I'_l, I_l)
\end{equation}
where $\alpha$ and $\beta$ are weighting coefficients.

The tampering mask $M$ is obtained by comparing the residual between the predefined location map $I_l$ and the reconstructed location map $I'_l$. We define the residual as:
\begin{equation}
 \text{M}[i, j] = \theta_{\tau}(\left| I_l[i, j] - I'_l[i, j] \right|)    
\end{equation}
where $\theta_{\tau}(x) = 1 \text{ if } x \geq \tau, \text{ otherwise } 0$. In our implementation, we set $\tau$ to $0.2$ based on our experience and observation.

\subsection{Intent Analysis and Interpretation}

The intent analysis and interpretation module is designed to interpret the manipulations identified by the watermark. This module leverages MLLMs to process detected tampering masks and infer the types of manipulations and their potential misleading effects. 

To leverage the spatial reasoning capabilities of MLLMs while preserving chart context, we introduce a visual prompting mechanism. Instead of using a dense binary mask $M$, we first compute the connected components from the damaged location map. The boundary of these components are then overlaid as outlines on the tampered image. This serves as a visual prompt for the MLLM. 

Our two-agent MLLM pipeline is designed to scaffold the user's analytical reasoning process based on bottom-up sensemaking process~\cite{pirolli2005sensemaking}. Specifically, our Component-to-Method Mapping rules (Fig.~\ref{fig:component_mapping}) are grounded in established visualization task taxonomies. These rules move the analysis beyond low-level pixel detection to a higher-level semantic inquiry~\cite{6634168}.
The Mask Refinement Agent automates the low-level perceptual task of ``anomaly identification,'' freeing the user's cognitive resources. The Intent Analysis Agent then addresses the higher-level analytical task of understanding why the anomaly exists.
The prompt structure, as shown in Fig.~\ref{fig:intentanalysisprompt}, consists of five components. The complete prompts are available in the appendix.

\subsubsection{Mask Refinement Agent}
The first agent, the Mask Refinement Agent, has two primary responsibilities. First, it refines the initial visual prompts to filter out noise. This is achieved by applying three analytical principles.

\textit{Area Analysis:} 
It prioritizes larger bounding boxes corresponding to significant chart elements (e.g., data points, axes) while discarding small, isolated ones in blank spaces that are likely noise.

\begin{figure}[tb]
\centering
\setlength{\fboxrule}{0.4pt}
\setlength{\fboxsep}{0cm}
\includegraphics[width=.99\linewidth]{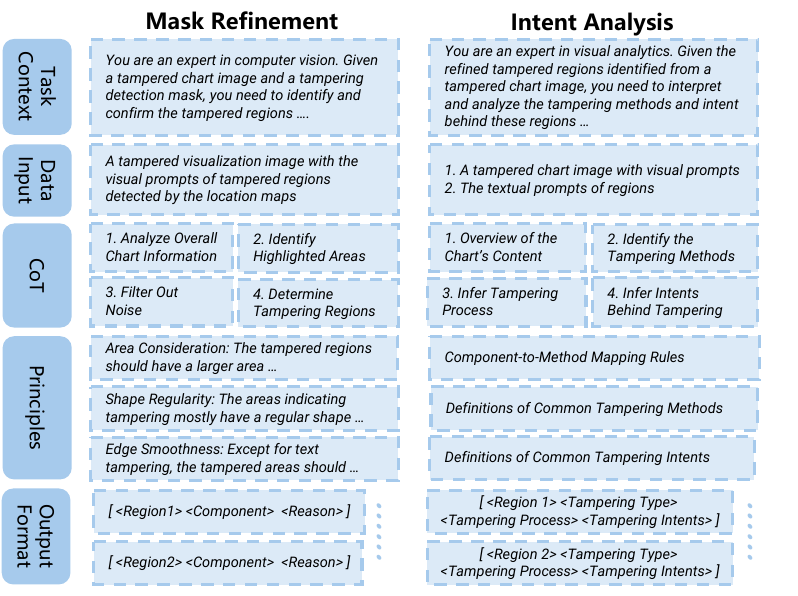}\ 

    \vspace{-5px}
\caption{\label{fig:intentanalysisprompt} Prompt structure of Intent Analysis and Interpretation Module.}

    \vspace{-5px}
\end{figure}

\textit{Shape Analysis:} 
It favors regular shapes (rectangles, circles) that align with structured chart elements, helping to distinguish intentional tampering from random watermark damage of data transmission.

\textit{Edge Analysis:} 
It checks for smooth, continuous edges in the highlighted regions, as irregular or jagged boundaries often indicate noise rather than a coherently manipulated element.

Second, the agent identifies the specific visualization component contained within each confirmed tampered region. Based on a predefined list, it categorizes each region as an \textit{axis}, \textit{data label}, \textit{legend}, \textit{colormap}, \textit{region} (for data elements like bars or points), \textit{logo}, or \textit{annotation}. The output of this agent is a set of refined tampering regions paired with their corresponding component labels, which serves as a structured input for the next agent.

\subsubsection{Intent Analysis Agent}
\begin{figure}[tb]
    \centering
    \includegraphics[width=\linewidth]{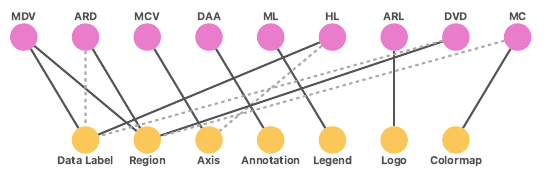}
    
    \vspace{-5px}
    \caption{The Component-to-Method mapping rules that guide the Intent Analysis Agent. Solid lines indicate a primary relationship (high likelihood), while dashed lines denote a secondary relationship. Tampering method abbreviations (top row) are defined in Fig.~\ref{fig:tamperType}.}
    \label{fig:component_mapping}
    
    \vspace{-10px}
\end{figure}

The Intent Analysis Agent is tasked with inferring the specific tampering method and the manipulator's underlying intent. To mitigate MLLM hallucination and ground its reasoning, this agent employs a constrained inference mechanism based on a predefined set of rules, as illustrated in Fig.~\ref{fig:component_mapping}. We defined these rules by mapping the tampering methods to the chart components they logically affect based on the existing misleading visualization gallery~\cite{lan2024junk} and benchmark~\cite{chen2025unmasking}.

This mechanism works via a Component-to-Method Mapping. The component label provided by the first agent is used to query the rule set, which provides a prioritized list of likely tampering methods. For instance, if the identified component is a \textit{data label}, the agent is instructed to first evaluate primary, high-likelihood methods such as ``Modifying data point values'' (MDV) or ``Hiding labels'' (HL). Only if these fail to explain the visual evidence will it consider secondary possibilities like ``Adding or removing data points'' (ARD). 

The agent's step-by-step reasoning process is as follows:

\textit{Infer Tampering Method:} For each tampered region, the agent selects the most appropriate tampering method by following the Component-to-Method Mapping rules (Fig.~\ref{fig:component_mapping}). The taxonomy of methods is derived from our formative study (Fig.~\ref{fig:tamperType}).

\textit{Describe Tampering Process:} The agent provides a concise, one-sentence description of how the manipulation was likely performed.

\textit{Infer Tampering Intent:} Finally, by analyzing the chosen method in the context of the overall visualization, the agent infers the potential motive behind the tampering.

\section{Application}

\begin{figure}[tbp]

    \centering
\includegraphics[width=\linewidth]{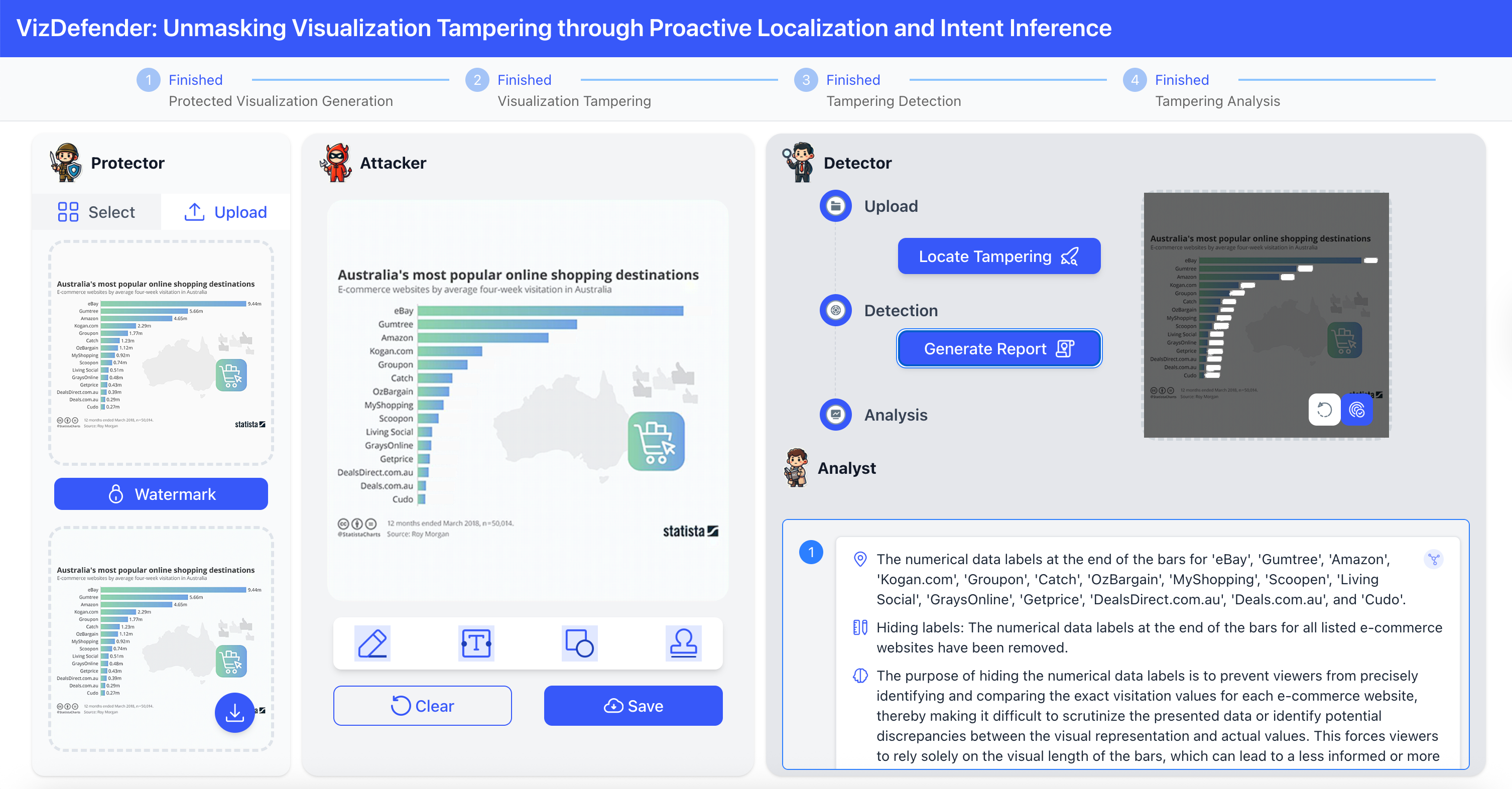}
\caption{\label{fig:system_ui} The web-based interface of our VizDefender system, demonstrating the end-to-end workflow from proactive protection and simulated tampering to automated detection and intent analysis.}
\vspace{-10px}

\end{figure}

VizDefender is designed as a solution for social media platforms. To demonstrate the practical workflow of our framework, we developed an interactive system that embodies the entire proactive defense lifecycle, as shown in Fig.~\ref{fig:system_ui}. The system guides different users through four key stages: (1) \textbf{Protector:} A content publisher uploads an original visualization. The social media platform watermarks the original visualization and outputs the protected visualization online. (2) \textbf{Attacker:} A hacker maliciously modifies the protected visualization and spreads the tampered visualization. (3) \textbf{Detector:} A content moderator of social media platform uploads the suspicious image to locate tampering. (4) \textbf{Analyst:} A fact-checker of social media platform or a curious user on social media reviews an automatically generated report detailing the manipulation and its inferred intent. This end-to-end process addresses the challenges of visual misinformation from two critical perspectives: scalable, platform-side content moderation and transparent, user-centric fact-checking.

\subsection{Automated Content Moderation at Scale}
Content moderation teams on social media platforms face an overwhelming, high-velocity stream of visual information, making manual review of every chart infeasible and leaving them vulnerable to sophisticated, large-scale misinformation campaigns. VizDefender addresses this critical need by functioning as an automated, platform-wide sentinel. Through a backend process of automatically watermarking all uploaded visualizations, it creates a fully verifiable ecosystem. This allows the platform's integrity systems to scan millions of images at scale and instantly flag any tampered content, such as the chart with subtly inflated data points in Fig.~\ref{fig:teaser}(C). More importantly, this automated process moves platforms beyond simple, reactive moderation. By aggregating detection data—common topics, recurring manipulation methods, or coordinated timing—platforms can gain strategic intelligence, identifying the macro-trends of disinformation and the potential organized efforts seeking to manipulate public discourse.

\subsection{Empowering Users with Critical Thinking Tools}
Beyond platform-level control, today's social media users are increasingly seeking tools that empower their own critical thinking. For many, simply knowing a visualization is ``fake'' is insufficient; they are curious about the manipulator's motive and want to understand \textit{how} and \textit{why} they are being misled. VizDefender directly meets this demand with a user-facing tool that emphasizes transparency and explanation. When a user activates the tool on a suspicious image, like the poverty map in Fig.~\ref{fig:teaser}(A). The system provides the evidence—the tampering mask revealing the manipulated colormap—and then addresses the crucial question of intent. The analysis module explains that the likely motive was to exaggerate the severity of poverty to evoke a stronger emotional response. This transforms a simple verification check into a learning moment, fostering a more discerning user base by revealing not just \textit{what} was faked, but the narrative purpose behind the deception.

\section{Evaluation}
We evaluate our method from three aspects: quality of protected visualization, accuracy of tampering detection, and accuracy of intent analysis. We conduct our experiments on a PC with a server with $4\times$NVIDIA RTX 3090. Our model is impemented with Pytorch. The weights (Equation~\ref{equ:total_loss}) are set as $\alpha=100$ and $\beta=1$ to balance the image quality of protected visualization and location map decoding.

\subsection{Test Dataset Composition}
Our test dataset is comprised of two main components: a manually created dataset and an automatically generated dataset.

\textbf{Manually Created Dataset (MCD)} was developed by recruiting five volunteers who are proficient in various image editing technologies, and possess a background in visualization design. Each volunteer was tasked with modifying $20$ visualizations according to nine distinct tampering types identified in our formative study (Fig.~\ref{fig:tamperType}), as well as creating examples with a mixture subset including at least two methods. The collection results in a total of $100$ visualizations. These volunteers were compensated at a rate of \$30 per hour for their time. The source visualizations for this dataset were sourced from Our World in Data~\cite{ourworlddata} and InfoGraphicVQA~\cite{Mathew_2022_WACV}, ensuring that there is no overlap with our training set. Volunteers were allowed to use any software they were familiar with to alter the visualization images. For each modified visualization, they saved a binary mask of the tampered regions as the ground truth. Additionally, they documented the tampering method and the intent behind each manipulation. To ensure the reliability of the data, two authors checked the dataset in a back-to-back manner.

\textbf{Automatically Generated Dataset (AGD)} was designed to evaluate the robustness of our method against various scales of tampering, without considering the intent behind the modifications. Due to the limitations of automated tampering methods, we focused on three primary tampering techniques that can be performed automatically: textual tampering, graphical element tampering, and painting.

\textbullet{} \textit{Textual Tampering:} We employed OCR~\cite{tesseractOCR} to extract text from the visualizations and used GPT-4o-mini to infer the semantics of the extracted text within the context of the visualization. The original text was then replaced with synonyms or strings with similar parts of speech, using similar fonts and sizes, or was outright deleted.

\textbullet{} \textit{Graphical Element Tampering:} Using the Segment Anything model~\cite{Kirillov_2023_ICCV}, we segmented and filtered graphical elements within the visualization. We then performed random deletions, modifications, or copy-paste actions on these elements.

\textbullet{} \textit{Painting:} Random circles, rectangles, and lines were added to the visualizations to simulate deceptive annotations.

The charts used for this dataset were sourced from the ChartQA-MLLM~\cite{10670526} dataset. For each tampering method, we created subsets with one, two, and three tampering instances per chart, resulting in $100$ charts per subset. This process was repeated for each of the three tampering methods, creating a total of $9$ subsets and $900$ charts.

\subsection{Image Quality of Protected Visualization}

\begin{table}[tbp]
\centering
\caption{\label{tab_eva:1}The image quality evaluation results of protected visualizations}
\vspace{-5px}

\small
\begin{tabular}{|c|ccc|ccc|}
\hline
\multirow{2}{*}{Method} & \multicolumn{3}{c|}{MCD}                                          & \multicolumn{3}{c|}{AGD}                            \\ \cline{2-7} 
                        & \multicolumn{1}{c|}{P $\uparrow$}     & \multicolumn{1}{c|}{S $\uparrow$}      & L $\downarrow$      & \multicolumn{1}{c|}{P $\uparrow$} & \multicolumn{1}{c|}{S $\uparrow$} & L $\downarrow$ \\ \hline
EditGuard               & \multicolumn{1}{c|}{32.57} & \multicolumn{1}{c|}{0.85} & 0.0057 & \multicolumn{1}{c|}{31.91}  & \multicolumn{1}{c|}{0.83}  & \multicolumn{1}{c|}{0.0114}  \\ \hline
\textbf{VizDefender}             & \multicolumn{1}{c|}{\textbf{33.55}} & \multicolumn{1}{c|}{\textbf{0.86}} & \textbf{0.0028} & \multicolumn{1}{c|}{\textbf{33.49}}  & \multicolumn{1}{c|}{\textbf{0.85}}  &  \multicolumn{1}{c|}{\textbf{0.0047}} \\ \hline
\end{tabular}

\vspace{-5px}
\end{table}

\begin{table}[tbp]
\centering
\caption{\label{tab_eva:2}The evaluation results of tampering detection on MCD}

\vspace{-5 px}
\renewcommand{\arraystretch}{1.1} 
\small
\begin{tabular}{|c|cc|cc|}
\hline
\multirow{2}{*}{Method} & \multicolumn{2}{c|}{No Tampering} & \multicolumn{2}{c|}{Post-Tampering} \\ \cline{2-5}
& \multicolumn{1}{c|}{Noise Per. $\downarrow$} & RMSE $\downarrow$ & \multicolumn{1}{c|}{IoU $\uparrow$} & F1 Score $\uparrow$ \\ \hline
ManTraNet & \multicolumn{1}{c|}{2.41\%} & 0.1466 & \multicolumn{1}{c|}{0.0539} & 0.0870 \\ \cline{1-5}
MVSS-Net & \multicolumn{1}{c|}{4.14\%} & 0.1628 & \multicolumn{1}{c|}{0.0182} & 0.0462 \\ \hline
EditGuard & \multicolumn{1}{c|}{0.45\%} & 0.0624 & \multicolumn{1}{c|}{0.5454} & 0.6669 \\ \cline{1-5}
\textbf{VizDefender} & \multicolumn{1}{c|}{\textbf{0.07\%}} & \textbf{0.0231} & \multicolumn{1}{c|}{\textbf{0.7272}} & \textbf{0.8259} \\ \hline
\end{tabular}

\vspace{-10px}
\end{table}

\begin{figure}[tbp]
\centering
\setlength{\fboxrule}{0.4pt}
\setlength{\fboxsep}{0cm}
\includegraphics[width=.99\linewidth]{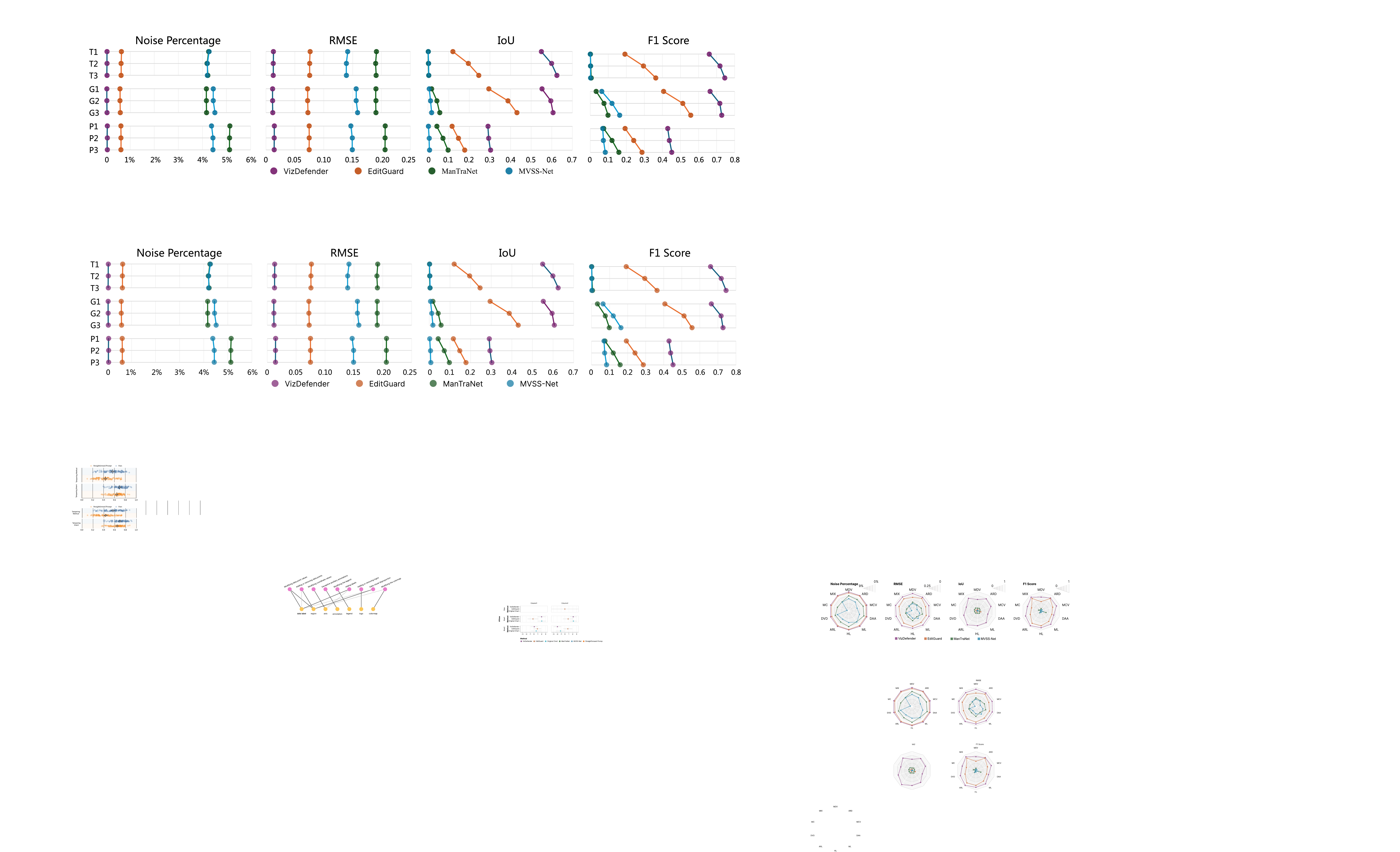}\
\caption{\label{fig:eva_detection_mcd} Comparison of tampering detection methods on the MCD across different tampering types. For Noise Percentage and RMSE, lower values (closer to the edge) are better, while for IoU and F1 Score, higher values (closer to the edge) are better. VizDefender demonstrates superior performance over all baselines across all tampering types and metrics.
}

\end{figure}

\begin{figure}[tb]
\centering
\setlength{\fboxrule}{0.4pt}
\setlength{\fboxsep}{0cm}
\includegraphics[width=.99\linewidth]{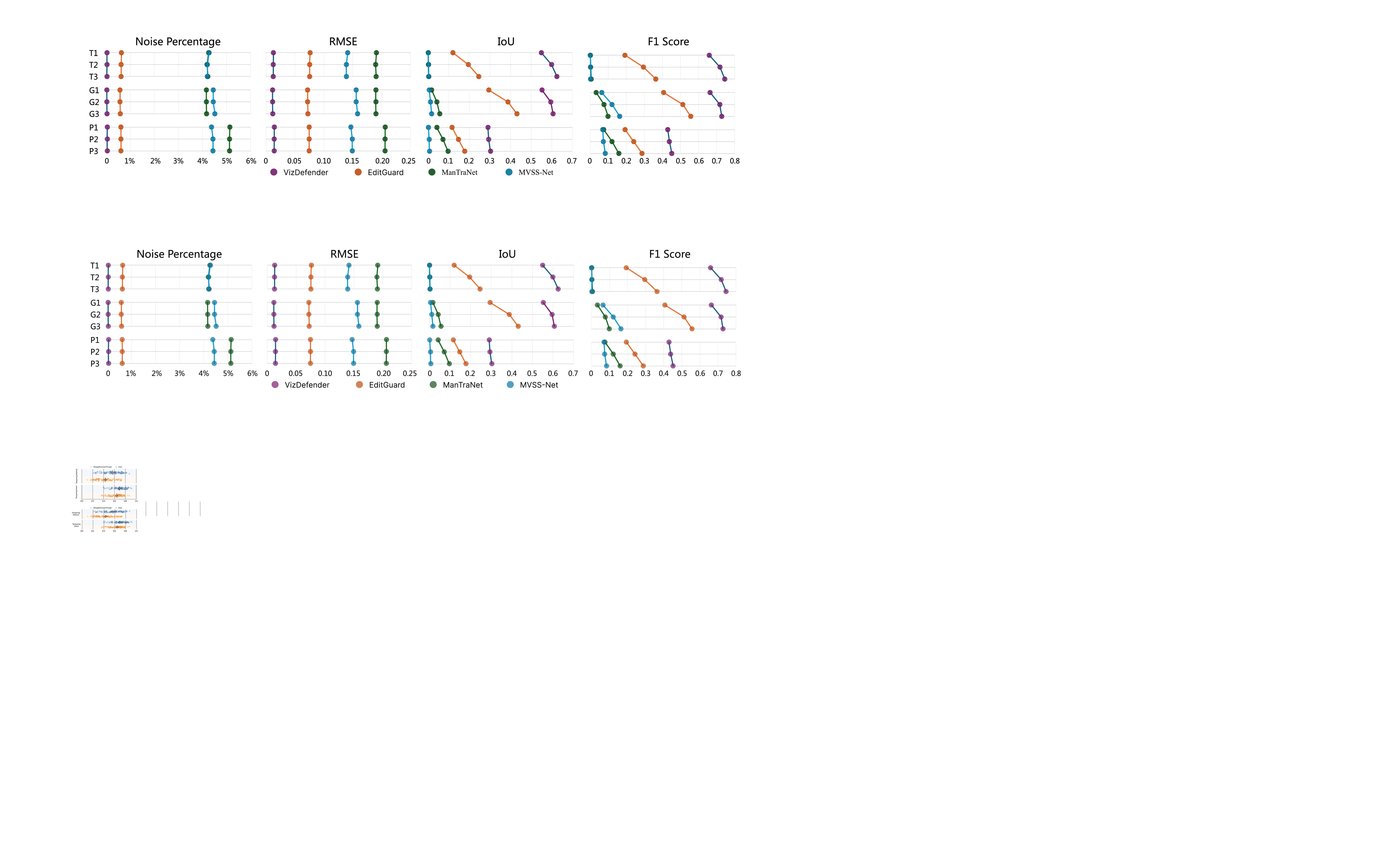}\ 
\caption{\label{fig:AGDeva} The evaluation results of tampering detection on the AGD. The dataset tests the robustness of different methods against various scales of tampering through three primary techniques: Textual Tampering (T1, T2, T3), Graphical Element Tampering (G1, G2, G3), and Painting (P1, P2, P3). Each technique is evaluated over subsets with one, two, and three instances of tampering per visualization. 
}

\vspace{-5px}
\end{figure}

The protected visualization is the visualization image with a semi-fragile watermark embedded into the input visualization through the encoding of localization maps. Our goal is to ensure that the protected visualization remains as visually similar to the original input visualization as possible. To evaluate the image quality of the protected visualization, we compare our method with the current state-of-the-art model for proactive defense against image tampering in natural images, EditGuard~\cite{zhang2024editguard}. We use three metrics for this comparison: Peak Signal-to-Noise Ratio (PSNR)~\cite{almohammad2010stego}, Structural Similarity Index (SSIM)~\cite{wang2004image}, and Learned Perceptual Image Patch Similarity (LPIPS)~\cite{zhang2018lpips}. PSNR measures the ratio between the maximum possible signal power and the power of corrupting noise, with higher values indicating better quality. SSIM evaluates the visual quality by comparing structural information, luminance, and contrast, where values closer to 1 indicate higher similarity. LPIPS assesses perceptual similarity based on deep features, with lower values indicating better similarity.

The results in Table~\ref{tab_eva:1} show that VizDefender outperforms EditGuard across all metrics on both MCD and AGD datasets. VizDefender achieves higher PSNR and SSIM values, indicating better visual fidelity and structural similarity to the original visualizations, while its lower LPIPS score reflects superior perceptual similarity. The differences are statistically significant ($p<0.001$) based on a Paired Sample t-Test.

\subsection{Accuracy of Tampering Detection}
To evaluate tampering detection accuracy, we compare VizDefender with two passive techniques, ManTraNet~\cite{wu2019mantra} and MVSS-Net~\cite{dong2022mvss}, and one proactive technique, EditGuard~\cite{zhang2024editguard}. Our evaluation uses four metrics and covers two scenarios: reliability on untampered images and sensitivity on tampered images. For reliability, we measure Noise Percentage and Root Mean Square Error (RMSE). For sensitivity, we use Intersection over Union (IoU) and the F1 Score to assess the accuracy of the predicted tampering masks against the ground truth.

The results on the MCD, summarized in Table~\ref{tab_eva:2}, demonstrate VizDefender's effectiveness. On untampered images, VizDefender shows superior reliability with the lowest noise ($0.07\%$) and RMSE ($0.0231$), indicating minimal false positives. On tampered images, it achieves the highest accuracy with an IoU of $0.7272$ and an F1 Score of $0.8259$. In contrast, the passive methods (ManTraNet, MVSS-Net) fail on this task, likely because they are designed for natural images. While EditGuard shows better generalization, it is still outperformed by our method, which is specifically tailored for visualizations.

This consistent superiority is further detailed in the per-category breakdown shown in Fig.~\ref{fig:eva_detection_mcd}.  The radar charts visualize performance across ten tampering types. These correspond to the nine distinct manipulation methods defined in Fig.~\ref{fig:tamperType}, plus a \textit{MIX} category representing a combination of two or more methods. The results show that VizDefender establishes a dominant performance margin in every scenario. It maintains the lowest error rates and highest detection scores regardless of the manipulation methods, from subtle data point value changes (MDV) to complex colormap modifications (MC). This highlights the robustness and versatility of our method against a diverse range of real-world tampering strategies.

The evaluation on the AGD, shown in Fig.~\ref{fig:AGDeva}, further confirms VizDefender's robustness. As the number of tampering instances increases (from 1 to 3), VizDefender maintains stable and high performance, while EditGuard's accuracy degrades more significantly. The passive methods fail entirely. While challenging operations like painting fine lines affect all methods, VizDefender consistently delivers the most accurate results. A detailed metric values and error analysis are available in the supplementary materials.

\subsection{Accuracy of Intent Inference}
We evaluate our Intent Analysis and Interpretation module by comparing it with a straightforward prompt-based baseline. The latter involves directly feeding the tampered visualization and detected location maps into an MLLM without the assistance of two specific agents. Our evaluation is conducted on the MCD, utilizing Gemini-Flash-2.5 as the MLLM. We assess performance using the categorical accuracy of the inferred tampering method, and the semantic similarity of the inferred tampering process and intent. For semantic similarity, we extract text embeddings using the Gemini-Text-Embedding model and calculate the cosine similarity between our model's outputs and the ground truth provided by the volunteers in our formative study.

\begin{figure}[tbp]
\centering
\setlength{\fboxrule}{0.4pt}
\setlength{\fboxsep}{0cm}
\includegraphics[width=.99\linewidth]{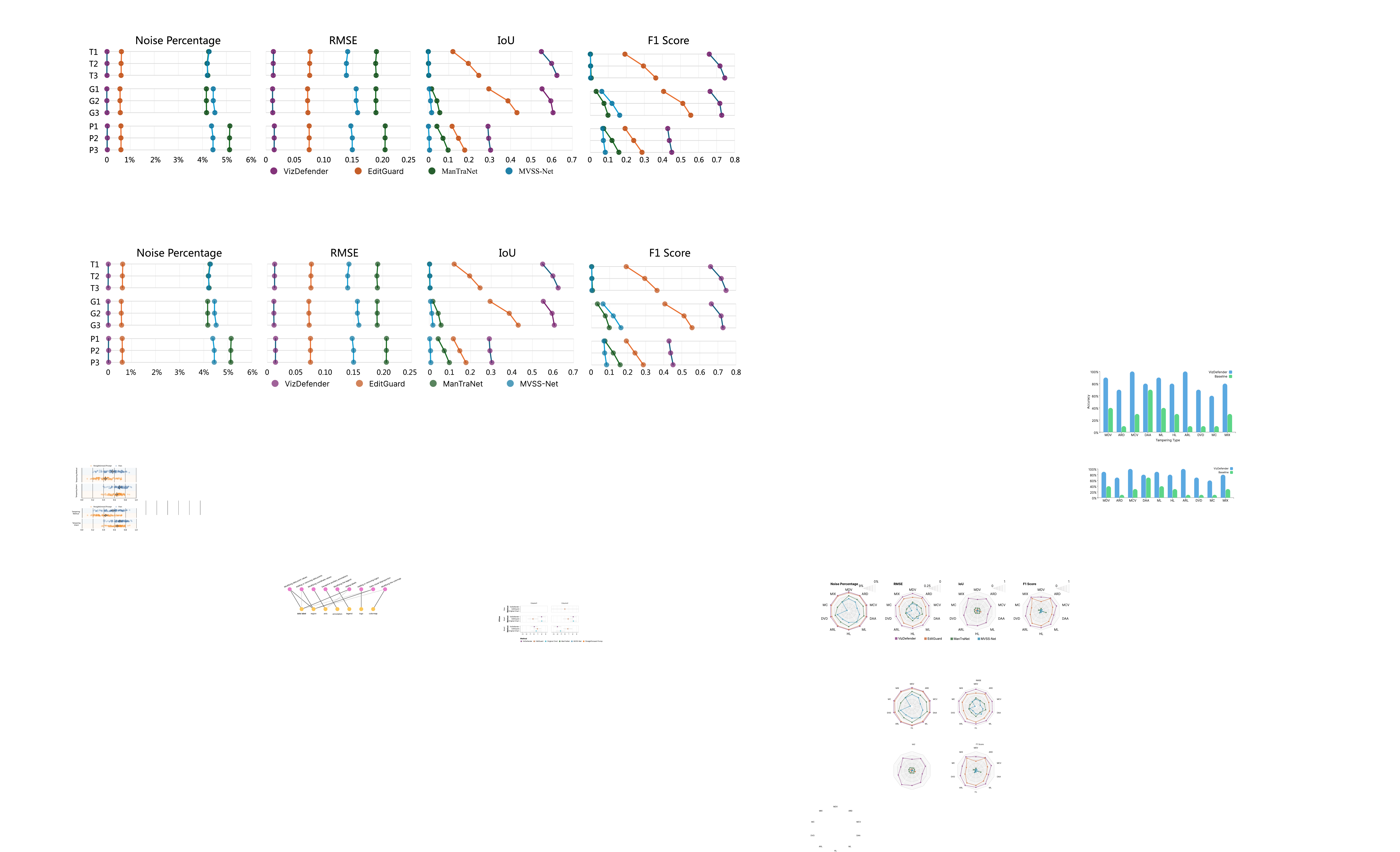}\
\caption{\label{fig:eva_tamperingtype}The evaluation results of tampering method identification comparing with the baseline. The X-axis represents different tampering type, and the Y-axis represents the accuracy.}
\end{figure}

First, we evaluate the model's ability to correctly identify the tampering method. We use a strict evaluation criterion: a classification is only considered correct if the set of identified tampering types perfectly matches the ground truth set. Any additional, incorrect, or omitted types render the identification incorrect. As shown in Fig.~\ref{fig:eva_tamperingtype}, VizDefender achieves a total classification accuracy of \textbf{82\%}, starkly outperforming the baseline's 28\%. This superiority is consistent across nearly all categories, and is particularly evident in cases like ``Modifying Coordinate Values'' (MCV) and ``Adding or Removing Logos'' (ARL), where VizDefender achieves 100\% accuracy compared to the baseline's 30\% and 10\%, respectively. This significant improvement in accuracy can be attributed to our structured Component-to-Method mapping, which constrains the MLLM's reasoning and prevents it from making ungrounded classifications--a common failure mode for the unconstrained baseline.

\begin{figure}[tbp]
\centering
\setlength{\fboxrule}{0.4pt}
\setlength{\fboxsep}{0cm}
\includegraphics[width=.99\linewidth]{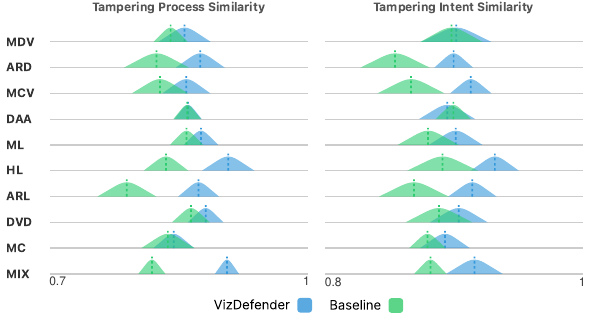}\
\caption{\label{fig:eva_intentanalysis} The evaluation results of tampering process and intent. The central line of the mountain represents the mean values, and the two ends represent the 95\% confidence interval.
}
\vspace{-10px}
\end{figure}

Next, we assess the semantic quality of the generated explanations for the tampering process and intent. The results, visualized as distributions in Fig.~\ref{fig:eva_intentanalysis}, show that VizDefender's outputs (blue) consistently achieve higher cosine similarity scores than the baseline (green) across almost all tampering types. On average, our method scores higher for both tampering process similarity (\textbf{0.874} vs. 0.836) and also for tampering intent similarity (\textbf{0.907} vs. 0.881). The tighter distributions for VizDefender also suggest a more stable and reliable performance.

The superior performance of our approach across both classification and semantic tasks is a direct result of our two-agent design. The Mask Refinement Agent provides clean and accurate spatial inputs, minimizing noise-induced errors. Subsequently, the Intent Analysis Agent, guided by its CoT framework and the crucial rule-based inference mechanism, drastically reduces model hallucination. 

Some relatively lower score for tampering method analysis can be attributed to the absence of prior knowledge about the original visualization, which the ground truth inherently possesses. For manipulations such as deletions or modifications of text, logos, or numerical values, inferring the tampering method solely from the tampered image and mask is challenging. However, our method remains effective at intent inference because it focuses on the narrative and contextual clues provided by the visualization's background story. More qualitative results and error analysis could be found in our supplementary materials.


\section{User Study}

We designed a user study to evaluate VizDefender's performance by a questionnaire. To ensure unbiased evaluation, all methods being compared were anonymized throughout the study, and the presentation order of different methods' results was randomized for each participant. Participants were not informed which results corresponded to which methods until after completing the study. The study was conducted through a 7-point Likert Scale consisting of three sections:

(1) \textit{Image Quality}: In this section, participants were presented with 5 sets of visualization images. Each set contained three versions of the same visualization: the original clean image, the image watermarked by VizDefender, and the image watermarked by EditGuard. Participants rated each image's quality and readability. 

(2) \textit{Tampering Detection}: This section showed participants 5 sets of tampered visualizations. Each set included the original visualization image, the tampered image, and tampering detection results from four different methods: VizDefender, EditGuard, MVSS-Net, and ManTraNet. Participants evaluated the accuracy of each method's detection mask. 

(3) \textit{Intent Inference}: The final section presented 5 tampering cases. For each case, participants were shown the tampered image with its detection mask, followed by two different tampering process and intent descriptions -- one predicted by straightforward prompting and another by VizDefender's Intent Analysis module. Participants rated the reasonableness of each output.

Before starting the evaluation, participants were briefed about the general purpose of visualization tampering detection and analysis. They were then guided through a practice example to familiarize themselves with the rating tasks. The actual study took about $20$ minutes.

\textbf{Participant:} We recruited 30 participants ($\mu_{age}=26.13$ years, $STD_{age}=2.70$, 15 males and 15 females) by distributing the study in university academic groups. All participants were graduate students or researchers. Their self-reported academic backgrounds confirmed relevant expertise: 30 participants were from diverse related fields including Data Visualization/Analytics (9), Computer Vision (8), Human-Computer Interaction (5), and Computer Graphics (2), Design and Art (2), Education (2), Psychology (1), and Journalism and Literature (1).

\begin{figure}[tbp]
    \centering
    \includegraphics[width=\linewidth]{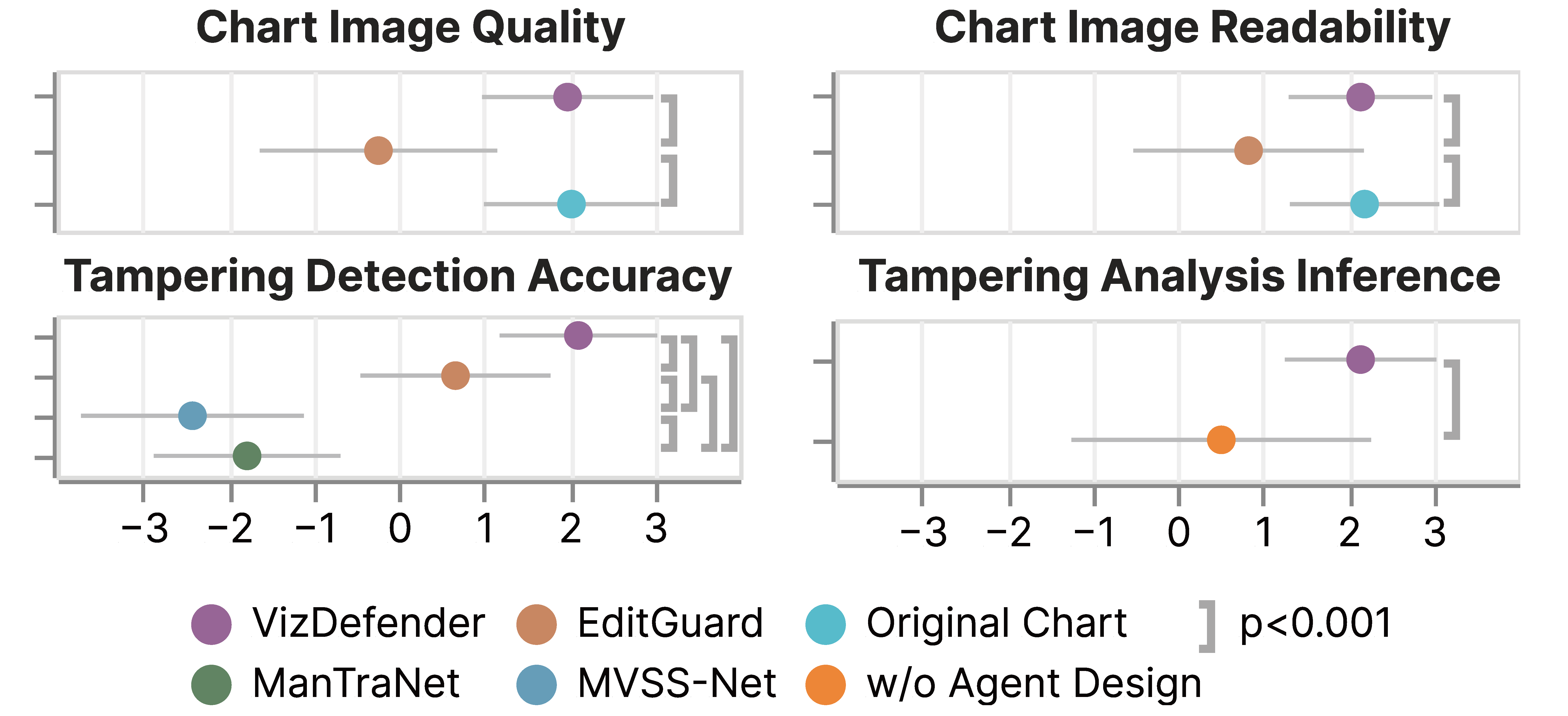}
    \caption{The results of our user study. Each circle represents the mean score, with error bars indicating the standard deviation. The statistical significance, was assessed using the Wilcoxon signed-rank test.}
    \label{fig:user_study}
    \vspace{-10px}
\end{figure}

\textbf{Analysis:} The results of our user study are as depicted in Fig.~\ref{fig:user_study}. Each evaluation metric in our user study was based on 150 samples, derived from 5 visualizations evaluated by 30 participants. The results indicate that our method excels in preserving image quality and readability, as users found it almost indistinguishable from the original visualizations. This highlights VizDefender's ability to integrate protective measures without compromising visual integrity. In contrast, EditGuard's watermarked images sometimes exhibited noticeable noise. This is likely due to the unique challenges posed by visualization images, which often contain large areas of blank space and continuous color blocks that are more susceptible to visual disruptions when watermarked. Our model, however, is specifically designed for visualization images, which enables it to manage these unique challenges while maintaining visual fidelity. 

In the tampering detection phase, VizDefender demonstrated superior performance (\textbf{2.113} vs. 0.673, -1.77, -2.41, p < 0.001) by achieving a low false positive rate and exhibiting sensitivity to subtle visualization manipulations, such as color changes in legends. User feedback indicates that our method can more accurately display tampered areas without requiring manual comparison. This provides the first crucial piece of evidence for supporting user reasoning: by accurately and automatically localizing the manipulation, our framework automates the low-level perceptual task and directs the user's attention.

In the intent inference analysis, VizDefender's Intent Analysis module also demonstrated advantages over the straightforward prompt-based approach (\textbf{2.15} vs. 0.52, p < 0.001). This can be linked to the module's utilization of agent-based techniques, which allows for a more nuanced understanding of the context and potential misleading effects of tampering. This proves how our framework supports a user's ability to reason. While the detection mask answers \textit{what} was changed, our module provides an answer for \textit{why} it was changed,. This high-level semantic interpretation moves beyond simple pixel-level detection and provides the user with the explicit context and motive needed to critically understand the impact of the manipulation, completing their reasoning process.

\section{Discussion}

\textbf{Redefining Tampering: The Sanctity of Semantic Integrity.} Detecting tampering in visualizations is fundamentally different from in natural images because visualizations are built on structured semantic encodings (e.g., axes, data points). We posit that this semantic integrity is paramount; therefore, any post-creation, image-level manipulation that alters the visual representation of data—even for seemingly benign rhetorical purposes—violates the original chart's integrity and authorial intent. This stance creates a unique technical challenge: the system must differentiate between semantically-irrelevant benign changes (e.g., compression) and semantically-relevant manipulations (e.g., altering an axis value). Our evaluations showed that VizDefender's domain-specific, semi-fragile approach is essential, as it is designed to protect not just pixels, but the semantic sanctity of the visualization. This highlights the necessity of specialized designs that protect not just pixels, but the semantic sanctity of the visualization.

\textbf{Rethinking Visual Prompts for MLLM-based Visualization Analysis.} 
A challenge we encountered was guiding the MLLM's focus onto specific tampered regions within a chart. While visual prompting is a maturing area for natural images~\cite{lin2025draw,zhang2024prompt}, our findings reveal that its application to data visualizations is fraught with unique difficulties. Standard prompting techniques, such as spotlights, masks, or highlighting, proved problematic as they inherently alter the chart's color encoding, thereby corrupting its original semantic meaning. For instance, a spotlight effect could mislead the MLLM into interpreting a bar's color as a different data category. Alternative methods also proved inadequate. Bounding boxes lack the required precision for dense visualizations; a single box drawn around a segment of a line in a multi-line chart often ambiguously encompasses multiple data series. Annotations using text or arrows risk being confused with the chart's native labels and graphical marks. Furthermore, we found that providing two separate images—one with and one without a visual prompt—led to poorer results. This suggests that current MLLMs possess weak spatial correspondence capabilities, and the alignment task between two images can introduce more hallucination and noise than it resolves.

Through trial-and-error, we used the contour lines of a tampered region's connected components as the visual prompt. This reveals a divergence between human and machine perception. While humans find such outlines subtle and less intuitive than a spotlight, the MLLM responded to this precise, non-destructive prompt with the highest accuracy. This led to a hybrid design in VizDefender: we use these contour lines as the backend prompt for the MLLM, while rendering a more human-friendly spotlight effect in the user-facing interface. This discovery underscores a broader implication for the field. It suggests that interacting with MLLMs on visualizations~\cite{yang2024askchart,fu2025refocus} may require a specialized visual language. Future research should focus on developing a dedicated visual prompting methodology for data visualizations—one that respects their semantic integrity and is tailored to the unique, and sometimes counter-intuitive, perceptual strengths of MLLMs, rather than simply adapting techniques designed for natural images.

\textbf{Trust in Visualizations Requires Proactive Defenses.} The ease with which visualization images can be tampered undermines public trust in data visualizations. The rapid dissemination of manipulated visual content on social media and other platforms underscores the limitations of reactive approaches to misinformation. Once tampered visualizations reach audiences, the damage to public perception is often irreversible, even if the manipulations are later exposed. Proactive defense systems address this challenge by embedding semi-fragile watermarks at the point of visualization creation, enabling tampering detection before manipulated visualizations can spread widely. Our proactive approach ensures that visualizations are protected from the outset, creating a layer of accountability for anyone attempting to manipulate them. Additionally, the integration of intent analysis allows stakeholders to assess the potential impact of tampering and respond accordingly. For social media platforms, our method provides an automated solution to identify and flag potentially manipulated visualizations during the content moderation process, helping prevent the viral spread of fake visualizations to maintain user trust.

\textbf{The Role of AI in Visual Misinformation: A Double-Edged Sword.} Our tools leverage AI to uncover subtle tampering and infer intent, empowering users to critically evaluate misinformation content.  However, AI is also a double-edged sword: while it enables advanced detection and analysis, generative AI technologies have lowered the barriers to creating convincing tampered visualizations, blurring the line between authentic and AI-generated content. The evolving landscape of AI-enabled misinformation emphasizes that technical solutions alone are insufficient. A comprehensive approach must combine technological defenses, with enhanced data literacy education to equip users with the skills to critically assess visualizations regardless of their source or creation method. This calls for a broader ecosystem of tools, policies, and educational initiatives to maintain trust in data visualization and safeguard its integrity in an era of rapidly advancing generative AI capabilities. As AI continues to evolve, the visualization community must adapt its strategies to preserve the trustworthiness of visual information and ensure its role as a reliable medium for communication.

\textbf{Limitation and Future Work.} 
\pvis{
The current version of VizDefender has several limitations in addressing visual misinformation detection and prevention. First, our system is unable to detect misinformation visualizations from scratch using generative AI tools. Such visualizations lack prior knowledge for comparison, posing a challenge for all existing tampering detection methods. Future work will explore statistical validation mechanisms to verify the consistency between raw data and its visual representation, as well as forensic techniques to identify inherent artifacts in AI-generated visualizations.
Second, the reliance on semi-fragile watermarking introduces vulnerabilities under adversarial attacks targeting watermark integrity. We plan to explore adaptive watermarking techniques and complementary detection methods.}\cl{SR-C(i),R2-Q8}
Third, the time performance of MLLMs is a limitation for real-time applications. On our MCD, the Mask Refinement Agent required an average of 11.01s (95\% CI: 10.57-11.45, T-test, p < 0.01) and the Intent Analysis Agent required 11.04s (95\% CI:10.49-11.60, T-test, p < 0.01). This latency presents a bottleneck for high-throughput moderation. Future work should explore optimization strategies, such as request parallelization and the use of smaller, specialized models.

Additionally, future directions include exploring interactive frameworks that allow users to steer the MLLM's reasoning based on their own mental model, as our current module provides a one-shot explanation. Transforming the tool into a collaborative reasoning partner is a key next step, along with integrating proactive defenses into popular visualization platforms, exploring blockchain-based verification for authentic visualizations, and conducting user studies to refine system personalization. We also plan to expand training and evaluation to these ``in-the-wild'' visualizations (e.g., infographics, news media charts) found on social media.

\section{conclusion}
In this paper, we addressed the critical challenge of malicious tampering in data visualizations, a growing threat to information integrity in the digital age. We introduced VizDefender, a proactive, end-to-end framework designed to not only detect manipulations but also to interpret their intent. Our approach combines a novel semi-fragile watermarking module for precise tampering localization with a sophisticated two-agent MLLM pipeline for intent analysis. This pipeline incorporates a constrained, rule-based reasoning mechanism that significantly reduces model hallucination and improves the accuracy of its interpretations.

Through extensive evaluations and user studies, we have demonstrated that VizDefender outperforms existing baselines in detection accuracy, visual quality, and the nuanced task of intent inference. Beyond its practical application as a tool for social media moderation, our work yields important insights for the broader AI and visualization communities. We acknowledge the limitations of our proactive approach, such as the inability to detect visualizations generated from scratch by AI and the inherent vulnerabilities of watermarking. Nonetheless, VizDefender represents a significant step toward safeguarding the trustworthiness of visual data. By providing a robust framework that bridges the gap between low-level pixel detection and high-level semantic interpretation, this work lays a critical foundation for future research in visualization security and helps ensure that data visualizations remain a reliable medium for communication.

\bibliographystyle{abbrv-doi-hyperref}

\bibliography{template}

\vfill

\newpage
\onecolumn
\appendix 

\section{Prompt Templates}
\label{prompts}

The following are the prompts for the mask refinement agent and the intent analysis agent in the intent analysis and interpretation module.

\begin{block}{Mask Refinement Agent}
\textit{\textbf{Task Context}}\\
\noindent\rule{\linewidth}{0.4pt}

You are an expert in computer vision. Given a tampered chart image, and we have detected the tampered areas and marked them with green lines.\\
You should provide the refined tampered regions and components as output, considering the principles provided.\\
The components to consider are:

\hspace{2em} - \texttt{axis}

\hspace{2em} - \texttt{data labels}

\hspace{2em} - \texttt{legend}

\hspace{2em} - \texttt{colormap}

\hspace{2em} - \texttt{region}

\hspace{2em} - \texttt{logo}

\hspace{2em} - \texttt{annotation}\\

\textit{\textbf{Data Input}}\\
\noindent\rule{\linewidth}{0.4pt}

    \texttt{Tampered Visualization Image with Visual Prmopt:} \highlight[orange]{$\{chart\_img\_path\}$} \\

\textit{\textbf{Chain-of-Thought}}\\
\noindent\rule{\linewidth}{0.4pt}

Your reasoning should retain only confirmed information. Here is a potential checklist for inferring the tampering area: \\
1. Analyze the overall information conveyed by the visualization. \\
2. Identify the areas surrounded by green lines, which indicate potential tampering regions. \\
3. Filter out noise from the highlighted areas and determine where meaningful tampering has actually occurred. \\
3. Determine the tampering regions and corresponding components. \\

\textit{\textbf{Principles}}\\
\noindent\rule{\linewidth}{0.4pt}

To filter out noise, you may refer to the three principles:\\

- \textbf{Area Analysis}: Prioritize highlighting sufficiently large areas and filter out minor areas along the background or edges that are likely noise. When you find a very small highlighted area on a relatively large color block, shape, or text, it is highly likely to be noise.\\

- \textbf{Shape Analysis}: For highlighted visual elements, focus on regular shapes (e.g., rectangles or circles) to detect potential tampering and disregard irregular shapes that suggest noise.\\

- \textbf{Edge Analysis}: Except for text modifications, the highlighted tampered areas should have smooth edges and fully cover the manipulated regions.\\

\textit{\textbf{Output Format}}\\
\noindent\rule{\linewidth}{0.4pt}

    Return the output in this strict format:\\
    \scriptsize
    \begin{verbatim}
     ```json
    {
      "tampered_regions": [
        {
          "tampered_region": "<tampered region>",
          "tampered_component": ["<tampered component 1>", "<tampered component 2>", ...]
          "reason": "reason for tampered region"
        },
        ...
      ]
    }
    '''
    \end{verbatim}
    \end{block}

\vfill
\newpage

\begin{block}{Intent Analysis Agent}
\small

\textit{\textbf{Task Context}}\\
\noindent\rule{\linewidth}{0.4pt}

You are an expert in visual analytics. Given the refined tampered regions identified from a tampered visualization image, you need to interpret and analyze the tampering intent behind these regions. \\
You should provide a detailed analysis 
of the tampering intent, including the types of manipulations and their potential misleading effects.
\\

\textit{\textbf{Data Input}}\\
\noindent\rule{\linewidth}{0.4pt}

    \texttt{Tampered Visualization Image with Visual Prompt:} \highlight[orange]{$\{chart\_img\_path\}$} \\
    \texttt{Textual Prompt of Tampered Region:} \highlight[orange]{$\{tampered\_region\_json\_path\}$} \\
    
\textit{\textbf{Component-to-Method Mapping Rules}}\\
\noindent\rule{\linewidth}{0.4pt}

For a given tampering\_component, you must For a given \texttt{tampering\_component}, you must evaluate the Primary Methods first. Only if none of the primary methods accurately describe the manipulation should you consider the Secondary Methods.

- If \texttt{tampering\_component} is \textbf{"region"}:\

\qquad - \textbf{Primary Methods:} \texttt{Modifying data point values}, \texttt{Adding or removing data points}, \texttt{Data-visual disproportion}\

\qquad - \textbf{Secondary Methods:} \texttt{Modifying the colormap}\

- If \texttt{tampering\_component} is \textbf{"data labels"}:\

\qquad - \textbf{Primary Methods:} \texttt{Modifying data point values}, \texttt{Hiding labels}\

\qquad - \textbf{Secondary Methods:} \texttt{Adding or removing data points}, \texttt{Data-visual disproportion}\

- If \texttt{tampering\_component} is \textbf{"axis"}:\

\qquad - \textbf{Primary Methods:} \texttt{Modifying coordinate values}\

\qquad - \textbf{Secondary Methods:} \texttt{Hiding labels}\

- If \texttt{tampering\_component} is \textbf{"legend"}: 

\qquad - \textbf{Primary Method:} \texttt{Modifying the legend}\

- If \texttt{tampering\_component} is \textbf{"annotation"}: 

\qquad - \textbf{Primary Method:} \texttt{Deceptive auxiliary annotations}\

- If \texttt{tampering\_component} is \textbf{"logo"}: 

\qquad - \textbf{Primary Method:} \texttt{Adding or removing logos}\

- If \texttt{tampering\_component} is \textbf{"colormap"}: 

\qquad - \textbf{Primary Method:} \texttt{Modifying the colormap}\\

\textit{\textbf{Chain-of-Thought}}\\
\noindent\rule{\linewidth}{0.4pt}

Let’s think step by step about the tampering intent. Here is a potential checklist for inferring the tampering intent:\\

1. Understand the context and a full understanding of the input\\
2. For each tampered region, first identify its \texttt{tampering\_component} from the input. Then, using the \textbf{Component-to-Method Mapping Rules} above, select the most fitting \texttt{method}. Remember to check Primary Methods before Secondary ones.\\
3. Based on your analysis, provide a simple description of the tampering process (\texttt{tamper}) in one sentence — how the original image was tampered to become the current version. For example: "Increase the population value of England from 53 million to 60 million and decrease the population value of Scotland from 5.2 million to 4 million.", "Remove the data points for feeling hopeful about the future" \\
4. Infer the \texttt{intent} based on the primary message conveyed by the chart and the identified tampered areas.\\

\textit{\textbf{Principles}}\\
\noindent\rule{\linewidth}{0.4pt}

To infer the tampering intents, you may refer to these common tampering types:\\
- \textbf{Modifying data point values}: Changing the quantitative value of an existing data point, which alters its visual representation (e.g., making a bar taller/shorter, moving a point up/down, changing the height of a line segment, or altering the boundary of an area). Crucially, in this method, the visual element is consistently updated to accurately reflect the new (modified) data point value. The visual element representing the data point remains present on the chart, but its specific value has been changed.\\
- \textbf{Adding or removing data points}: Introducing entirely new data points (and their corresponding visual elements) or completely deleting existing data points (and their corresponding visual elements). This results in visual elements appearing or disappearing from the chart, rather than just changing their existing properties.\\
- \textbf{Modifying coordinate values}: Changing the values or textual labels of x-axis or y-axis to alter their position in the chart.\\
- \textbf{Deceptive auxiliary annotations}: Using misleading annotations (e.g., clustering boxes, guide lines, arrows) to create a false impression.\\
- \textbf{Modifying the legend}: Changing the legend’s content, colors, or order to mislead viewers about what the data represents.\\
- \textbf{Hiding labels}: Removing data labels near data points (e.g., scatter, point, etc.) to obscure the true meaning of the data.\\
- \textbf{Adding or removing logos}: Inserting or deleting logos to mislead viewers about the data source.\\
- \textbf{Data-visual disproportion}: Making the visual representation of data inconsistent with the actual values. This occurs when the visual element (e.g., bar length, area size) does not accurately correspond to its stated numerical value, or when one is modified without the other being consistently updated, creating a mismatch.\\
- \textbf{Modifying the colormap}: Adjusting the color mapping (including legend, data points, and their background colors) to distort the perception of data distribution, or introducing inconsistent colors to mislead.\\
- \textbf{Others}: Any other tampering method that is not included in the above types.\\

\textit{\textbf{Output Format}}\\
\noindent\rule{\linewidth}{0.4pt}

    Return in JSON format with the following structure:

    \vspace{-10px}
    \small
    \begin{verbatim}
     ```json
    { "tampering_intents": [
        {
          "tampered_region": "<tampered region>",
          "method": "<Tampering Method>",
          "tamper": "<Tampering Process>",
          "intent": "<Tampering Intent>"
        },...]}
    '''
    \end{verbatim}
    \end{block}

\vfill
\newpage

\section{Quantitative Evaluation Results}

\subsection{Image Quality of Protected Visualization}

\begin{table}[!h]
\renewcommand{\arraystretch}{1.4}
\centering
\caption{\label{supp_tab_1}The image quality evaluation results of protected visualizations on the MCD. The Mean column represents the average value of each metric (SSIM, PSNR, LPIPS). The Lower and Upper columns indicate the 95\% confidence interval for each metric.}

\footnotesize
\begin{tabular}{|c|l|ccc|ccc|ccc|}
\hline
\multirow{2}{*}{Method}       & \multicolumn{1}{c|}{\multirow{2}{*}{Subset}} & \multicolumn{3}{c|}{SSIM $\uparrow$}                                          & \multicolumn{3}{c|}{PSNR $\uparrow$}                                             & \multicolumn{3}{c|}{LPIPS $\downarrow$}                                         \\ \cline{3-11} 
                              & \multicolumn{1}{c|}{}                                & \multicolumn{1}{c|}{Mean}   & \multicolumn{1}{c|}{Lower}  & Upper  & \multicolumn{1}{c|}{Mean}    & \multicolumn{1}{c|}{Lower}   & Upper   & \multicolumn{1}{c|}{Mean}   & \multicolumn{1}{c|}{Lower}  & Upper  \\ \hline
\multirow{10}{*}{\textbf{VizDefender}} & Modifying Data Point Values                          & \multicolumn{1}{c|}{0.8576} & \multicolumn{1}{c|}{0.8436} & 0.8716 & \multicolumn{1}{c|}{33.5409} & \multicolumn{1}{c|}{33.2815} & 33.8003 & \multicolumn{1}{c|}{0.0030} & \multicolumn{1}{c|}{0.0022} & 0.0039 \\ \cline{2-11} 
                              & Adding or Removing Data Points                       & \multicolumn{1}{c|}{0.8582} & \multicolumn{1}{c|}{0.8441} & 0.8722 & \multicolumn{1}{c|}{33.2812} & \multicolumn{1}{c|}{32.8249} & 33.7374 & \multicolumn{1}{c|}{0.0023} & \multicolumn{1}{c|}{0.0018} & 0.0028 \\ \cline{2-11} 
                              & Modifying Coordinate Values                          & \multicolumn{1}{c|}{0.8587} & \multicolumn{1}{c|}{0.8515} & 0.8658 & \multicolumn{1}{c|}{33.4550} & \multicolumn{1}{c|}{33.3407} & 33.5692 & \multicolumn{1}{c|}{0.0032} & \multicolumn{1}{c|}{0.0027} & 0.0037 \\ \cline{2-11} 
                              & Deceptive Auxiliary Annotations                      & \multicolumn{1}{c|}{0.8557} & \multicolumn{1}{c|}{0.8485} & 0.8628 & \multicolumn{1}{c|}{33.6791} & \multicolumn{1}{c|}{33.4246} & 33.9337 & \multicolumn{1}{c|}{0.0032} & \multicolumn{1}{c|}{0.0024} & 0.0040 \\ \cline{2-11} 
                              & Modifying Legend                                     & \multicolumn{1}{c|}{0.8575} & \multicolumn{1}{c|}{0.8487} & 0.8662 & \multicolumn{1}{c|}{33.3777} & \multicolumn{1}{c|}{33.1043} & 33.6512 & \multicolumn{1}{c|}{0.0031} & \multicolumn{1}{c|}{0.0026} & 0.0037 \\ \cline{2-11} 
                              & Hiding Labels                                        & \multicolumn{1}{c|}{0.8565} & \multicolumn{1}{c|}{0.8344} & 0.8785 & \multicolumn{1}{c|}{33.6103} & \multicolumn{1}{c|}{33.0719} & 34.1488 & \multicolumn{1}{c|}{0.0028} & \multicolumn{1}{c|}{0.0023} & 0.0032 \\ \cline{2-11} 
                              & Adding or Removing Logos                             & \multicolumn{1}{c|}{0.8515} & \multicolumn{1}{c|}{0.8377} & 0.8654 & \multicolumn{1}{c|}{33.5891} & \multicolumn{1}{c|}{33.1479} & 34.0303 & \multicolumn{1}{c|}{0.0027} & \multicolumn{1}{c|}{0.0022} & 0.0032 \\ \cline{2-11} 
                              & Data-Visual Disproportion                            & \multicolumn{1}{c|}{0.8857} & \multicolumn{1}{c|}{0.8752} & 0.8961 & \multicolumn{1}{c|}{34.0616} & \multicolumn{1}{c|}{33.8462} & 34.2770 & \multicolumn{1}{c|}{0.0018} & \multicolumn{1}{c|}{0.0016} & 0.0020 \\ \cline{2-11} 
                              & Modifying Colormap                                   & \multicolumn{1}{c|}{0.8620} & \multicolumn{1}{c|}{0.8446} & 0.8794 & \multicolumn{1}{c|}{33.5988} & \multicolumn{1}{c|}{33.2944} & 33.9031 & \multicolumn{1}{c|}{0.0026} & \multicolumn{1}{c|}{0.0020} & 0.0031 \\ \cline{2-11} 
                              & Mixture                                              & \multicolumn{1}{c|}{0.8633} & \multicolumn{1}{c|}{0.8520} & 0.8745 & \multicolumn{1}{c|}{33.3248} & \multicolumn{1}{c|}{32.6344} & 34.0152 & \multicolumn{1}{c|}{0.0028} & \multicolumn{1}{c|}{0.0022} & 0.0034 \\ \hline
\multirow{10}{*}{EditGuard}   & Modifying Data Point Values                          & \multicolumn{1}{c|}{0.8474} & \multicolumn{1}{c|}{0.8327} & 0.8621 & \multicolumn{1}{c|}{32.3916} & \multicolumn{1}{c|}{31.4849} & 33.2983 & \multicolumn{1}{c|}{0.0067} & \multicolumn{1}{c|}{0.0047} & 0.0088 \\ \cline{2-11} 
                              & Adding or Removing Data Points                       & \multicolumn{1}{c|}{0.8517} & \multicolumn{1}{c|}{0.8366} & 0.8668 & \multicolumn{1}{c|}{32.4421} & \multicolumn{1}{c|}{31.9697} & 32.9144 & \multicolumn{1}{c|}{0.0043} & \multicolumn{1}{c|}{0.0032} & 0.0054 \\ \cline{2-11} 
                              & Modifying Coordinate Values                          & \multicolumn{1}{c|}{0.8461} & \multicolumn{1}{c|}{0.8358} & 0.8564 & \multicolumn{1}{c|}{32.5298} & \multicolumn{1}{c|}{32.1043} & 32.9553 & \multicolumn{1}{c|}{0.0057} & \multicolumn{1}{c|}{0.0046} & 0.0068 \\ \cline{2-11} 
                              & Deceptive Auxiliary Annotations                      & \multicolumn{1}{c|}{0.8427} & \multicolumn{1}{c|}{0.8289} & 0.8566 & \multicolumn{1}{c|}{33.0031} & \multicolumn{1}{c|}{32.6574} & 33.3488 & \multicolumn{1}{c|}{0.0057} & \multicolumn{1}{c|}{0.0046} & 0.0069 \\ \cline{2-11} 
                              & Modifying Legend                                     & \multicolumn{1}{c|}{0.8483} & \multicolumn{1}{c|}{0.8354} & 0.8612 & \multicolumn{1}{c|}{32.7114} & \multicolumn{1}{c|}{32.3867} & 33.0361 & \multicolumn{1}{c|}{0.0057} & \multicolumn{1}{c|}{0.0046} & 0.0067 \\ \cline{2-11} 
                              & Hiding Labels                                        & \multicolumn{1}{c|}{0.8482} & \multicolumn{1}{c|}{0.8243} & 0.8721 & \multicolumn{1}{c|}{32.4081} & \multicolumn{1}{c|}{31.8434} & 32.9728 & \multicolumn{1}{c|}{0.0057} & \multicolumn{1}{c|}{0.0047} & 0.0067 \\ \cline{2-11} 
                              & Adding or Removing Logos                             & \multicolumn{1}{c|}{0.8461} & \multicolumn{1}{c|}{0.8329} & 0.8593 & \multicolumn{1}{c|}{32.8864} & \multicolumn{1}{c|}{32.3671} & 33.4056 & \multicolumn{1}{c|}{0.0053} & \multicolumn{1}{c|}{0.0036} & 0.0070 \\ \cline{2-11} 
                              & Data-Visual Disproportion                            & \multicolumn{1}{c|}{0.8776} & \multicolumn{1}{c|}{0.8667} & 0.8884 & \multicolumn{1}{c|}{32.8792} & \multicolumn{1}{c|}{32.5107} & 33.2477 & \multicolumn{1}{c|}{0.0051} & \multicolumn{1}{c|}{0.0039} & 0.0063 \\ \cline{2-11} 
                              & Modifying Colormap                                   & \multicolumn{1}{c|}{0.8550} & \multicolumn{1}{c|}{0.8372} & 0.8727 & \multicolumn{1}{c|}{32.4965} & \multicolumn{1}{c|}{31.9868} & 33.0063 & \multicolumn{1}{c|}{0.0063} & \multicolumn{1}{c|}{0.0037} & 0.0089 \\ \cline{2-11} 
                              & Mixture                                              & \multicolumn{1}{c|}{0.8550} & \multicolumn{1}{c|}{0.8441} & 0.8658 & \multicolumn{1}{c|}{31.9244} & \multicolumn{1}{c|}{30.7683} & 33.0805 & \multicolumn{1}{c|}{0.0069} & \multicolumn{1}{c|}{0.0042} & 0.0095 \\ \hline
\end{tabular}
\end{table}

\begin{table}[!h]
\renewcommand{\arraystretch}{1.4}
\centering
\caption{\label{supp_tab_2}The image quality evaluation results of protected visualizations on the AGD. The Mean column represents the average value of each metric (SSIM, PSNR, LPIPS). The Lower and Upper columns indicate the 95\% confidence interval for each metric.}

\footnotesize
\begin{tabular}{|c|l|ccc|ccc|ccc|}
\hline
\multirow{2}{*}{Method}      & \multicolumn{1}{c|}{\multirow{2}{*}{Subset}} & \multicolumn{3}{c|}{SSIM $\uparrow$}                                          & \multicolumn{3}{c|}{PSNR $\uparrow$}                                             & \multicolumn{3}{c|}{LPIPS $\downarrow$}                                         \\ \cline{3-11} 
                             & \multicolumn{1}{c|}{}                        & \multicolumn{1}{c|}{Mean}   & \multicolumn{1}{c|}{Lower}  & Upper  & \multicolumn{1}{c|}{Mean}    & \multicolumn{1}{c|}{Lower}   & Upper   & \multicolumn{1}{c|}{Mean}   & \multicolumn{1}{c|}{Lower}  & Upper  \\ \hline
\multirow{9}{*}{\textbf{VizDefender}} & Textual Tampering - 1                        & \multicolumn{1}{c|}{0.8513} & \multicolumn{1}{c|}{0.8476} & 0.8549 & \multicolumn{1}{c|}{33.4918} & \multicolumn{1}{c|}{33.2611} & 33.7225 & \multicolumn{1}{c|}{0.0046} & \multicolumn{1}{c|}{0.0040} & 0.0052 \\ \cline{2-11} 
                             & Textual Tampering - 2                        & \multicolumn{1}{c|}{0.8511} & \multicolumn{1}{c|}{0.8474} & 0.8549 & \multicolumn{1}{c|}{33.5303} & \multicolumn{1}{c|}{33.3143} & 33.7463 & \multicolumn{1}{c|}{0.0046} & \multicolumn{1}{c|}{0.0040} & 0.0052 \\ \cline{2-11} 
                             & Textual Tampering - 3                        & \multicolumn{1}{c|}{0.8509} & \multicolumn{1}{c|}{0.8472} & 0.8546 & \multicolumn{1}{c|}{33.4781} & \multicolumn{1}{c|}{33.2484} & 33.7079 & \multicolumn{1}{c|}{0.0047} & \multicolumn{1}{c|}{0.0041} & 0.0053 \\ \cline{2-11} 
                             & Graphical Tampering - 1                      & \multicolumn{1}{c|}{0.8487} & \multicolumn{1}{c|}{0.8448} & 0.8527 & \multicolumn{1}{c|}{33.4588} & \multicolumn{1}{c|}{33.2744} & 33.6431 & \multicolumn{1}{c|}{0.0047} & \multicolumn{1}{c|}{0.0042} & 0.0052 \\ \cline{2-11} 
                             & Graphical Tampering - 2                      & \multicolumn{1}{c|}{0.8487} & \multicolumn{1}{c|}{0.8448} & 0.8527 & \multicolumn{1}{c|}{33.4588} & \multicolumn{1}{c|}{33.2744} & 33.6431 & \multicolumn{1}{c|}{0.0047} & \multicolumn{1}{c|}{0.0042} & 0.0052 \\ \cline{2-11} 
                             & Graphical Tampering - 3                      & \multicolumn{1}{c|}{0.8491} & \multicolumn{1}{c|}{0.8451} & 0.8531 & \multicolumn{1}{c|}{33.4284} & \multicolumn{1}{c|}{33.2341} & 33.6228 & \multicolumn{1}{c|}{0.0047} & \multicolumn{1}{c|}{0.0042} & 0.0052 \\ \cline{2-11} 
                             & Painting - 1                                 & \multicolumn{1}{c|}{0.8513} & \multicolumn{1}{c|}{0.8474} & 0.8552 & \multicolumn{1}{c|}{33.5253} & \multicolumn{1}{c|}{33.3118} & 33.7389 & \multicolumn{1}{c|}{0.0048} & \multicolumn{1}{c|}{0.0041} & 0.0054 \\ \cline{2-11} 
                             & Painting - 2                                 & \multicolumn{1}{c|}{0.8513} & \multicolumn{1}{c|}{0.8474} & 0.8552 & \multicolumn{1}{c|}{33.5191} & \multicolumn{1}{c|}{33.3047} & 33.7335 & \multicolumn{1}{c|}{0.0048} & \multicolumn{1}{c|}{0.0041} & 0.0054 \\ \cline{2-11} 
                             & Painting - 3                                 & \multicolumn{1}{c|}{0.8512} & \multicolumn{1}{c|}{0.8473} & 0.8552 & \multicolumn{1}{c|}{33.5164} & \multicolumn{1}{c|}{33.3019} & 33.7309 & \multicolumn{1}{c|}{0.0048} & \multicolumn{1}{c|}{0.0042} & 0.0054 \\ \hline
\multirow{9}{*}{EditGuard}   & Textual Tampering - 1                        & \multicolumn{1}{c|}{0.8345} & \multicolumn{1}{c|}{0.8298} & 0.8392 & \multicolumn{1}{c|}{32.0800} & \multicolumn{1}{c|}{31.8087} & 32.3513 & \multicolumn{1}{c|}{0.0110} & \multicolumn{1}{c|}{0.0101} & 0.0120 \\ \cline{2-11} 
                             & Textual Tampering - 2                        & \multicolumn{1}{c|}{0.8340} & \multicolumn{1}{c|}{0.8293} & 0.8387 & \multicolumn{1}{c|}{32.0741} & \multicolumn{1}{c|}{31.8037} & 32.3446 & \multicolumn{1}{c|}{0.0111} & \multicolumn{1}{c|}{0.0101} & 0.0120 \\ \cline{2-11} 
                             & Textual Tampering - 3                        & \multicolumn{1}{c|}{0.8337} & \multicolumn{1}{c|}{0.8289} & 0.8384 & \multicolumn{1}{c|}{32.0629} & \multicolumn{1}{c|}{31.7912} & 32.3346 & \multicolumn{1}{c|}{0.0111} & \multicolumn{1}{c|}{0.0101} & 0.0121 \\ \cline{2-11} 
                             & Graphical Tampering - 1                      & \multicolumn{1}{c|}{0.8293} & \multicolumn{1}{c|}{0.8245} & 0.8340 & \multicolumn{1}{c|}{31.6900} & \multicolumn{1}{c|}{31.3810} & 31.9990 & \multicolumn{1}{c|}{0.0122} & \multicolumn{1}{c|}{0.0110} & 0.0134 \\ \cline{2-11} 
                             & Graphical Tampering - 2                      & \multicolumn{1}{c|}{0.8293} & \multicolumn{1}{c|}{0.8245} & 0.8340 & \multicolumn{1}{c|}{31.6900} & \multicolumn{1}{c|}{31.3810} & 31.9990 & \multicolumn{1}{c|}{0.0122} & \multicolumn{1}{c|}{0.0110} & 0.0134 \\ \cline{2-11} 
                             & Graphical Tampering - 3                      & \multicolumn{1}{c|}{0.8299} & \multicolumn{1}{c|}{0.8251} & 0.8347 & \multicolumn{1}{c|}{31.6523} & \multicolumn{1}{c|}{31.3389} & 31.9658 & \multicolumn{1}{c|}{0.0123} & \multicolumn{1}{c|}{0.0111} & 0.0135 \\ \cline{2-11} 
                             & Painting - 1                                 & \multicolumn{1}{c|}{0.8342} & \multicolumn{1}{c|}{0.8293} & 0.8391 & \multicolumn{1}{c|}{31.9776} & \multicolumn{1}{c|}{31.6498} & 32.3055 & \multicolumn{1}{c|}{0.0110} & \multicolumn{1}{c|}{0.0098} & 0.0122 \\ \cline{2-11} 
                             & Painting - 2                                 & \multicolumn{1}{c|}{0.8342} & \multicolumn{1}{c|}{0.8293} & 0.8392 & \multicolumn{1}{c|}{31.9607} & \multicolumn{1}{c|}{31.6314} & 32.2901 & \multicolumn{1}{c|}{0.0111} & \multicolumn{1}{c|}{0.0098} & 0.0123 \\ \cline{2-11} 
                             & Painting - 3                                 & \multicolumn{1}{c|}{0.8345} & \multicolumn{1}{c|}{0.8296} & 0.8393 & \multicolumn{1}{c|}{31.9600} & \multicolumn{1}{c|}{31.6307} & 32.2893 & \multicolumn{1}{c|}{0.0111} & \multicolumn{1}{c|}{0.0099} & 0.0123 \\ \hline
\end{tabular}
\end{table}

\vfill
\newpage

\subsection{Accuracy of Tampering Detection}

\begin{table}[!h]
\renewcommand{\arraystretch}{1.5}
\centering
\caption{\label{supp_tab_4}The accuracy evaluation results of tampering detection on the AGD. The Mean column represents the average value of each metric (Noise Percentage, RMSE, IoU, F1 Score). The Lower and Upper columns indicate the 95\% confidence interval for each metric.}

\fontsize{6.5}{7.8}\selectfont
\begin{tabular}{|c|l|cccccc|cccccc|}
\hline
\multirow{3}{*}{Method}       & \multicolumn{1}{c|}{\multirow{3}{*}{Subset}} & \multicolumn{6}{c|}{No Tampering}                                                                                                                              & \multicolumn{6}{c|}{Post-Tampering}                                                                                                                      \\ \cline{3-14} 
                              & \multicolumn{1}{c|}{}                        & \multicolumn{3}{c|}{Noise Percentage $\downarrow$}                                                        & \multicolumn{3}{c|}{RMSE $\downarrow$}                                       & \multicolumn{3}{c|}{IoU $\uparrow$}                                                              & \multicolumn{3}{c|}{F1 Score $\uparrow$}                                    \\ \cline{3-14} 
                              & \multicolumn{1}{c|}{}                        & \multicolumn{1}{c|}{Mean}    & \multicolumn{1}{c|}{Lower}    & \multicolumn{1}{c|}{Upper}    & \multicolumn{1}{c|}{Mean}  & \multicolumn{1}{c|}{Lower} & Upper & \multicolumn{1}{c|}{Mean}  & \multicolumn{1}{c|}{Lower}  & \multicolumn{1}{c|}{Upper} & \multicolumn{1}{c|}{Mean}  & \multicolumn{1}{c|}{Lower}  & Upper \\ \hline
\multirow{10}{*}{\textbf{VizDefender}} & Modifying Data Point Values                  & \multicolumn{1}{c|}{0.061\%} & \multicolumn{1}{c|}{0.032\%}  & \multicolumn{1}{c|}{0.090\%}  & \multicolumn{1}{c|}{0.023} & \multicolumn{1}{c|}{0.016} & 0.030 & \multicolumn{1}{c|}{0.590} & \multicolumn{1}{c|}{0.445}  & \multicolumn{1}{c|}{0.736} & \multicolumn{1}{c|}{0.722} & \multicolumn{1}{c|}{0.593}  & 0.850 \\ \cline{2-14} 
                              & Adding or Removing Data Points               & \multicolumn{1}{c|}{0.103\%} & \multicolumn{1}{c|}{0.045\%}  & \multicolumn{1}{c|}{0.161\%}  & \multicolumn{1}{c|}{0.030} & \multicolumn{1}{c|}{0.020} & 0.039 & \multicolumn{1}{c|}{0.771} & \multicolumn{1}{c|}{0.561}  & \multicolumn{1}{c|}{0.981} & \multicolumn{1}{c|}{0.834} & \multicolumn{1}{c|}{0.655}  & 1.012 \\ \cline{2-14} 
                              & Modifying Coordinate Values                  & \multicolumn{1}{c|}{0.102\%} & \multicolumn{1}{c|}{0.028\%}  & \multicolumn{1}{c|}{0.176\%}  & \multicolumn{1}{c|}{0.028} & \multicolumn{1}{c|}{0.016} & 0.039 & \multicolumn{1}{c|}{0.733} & \multicolumn{1}{c|}{0.661}  & \multicolumn{1}{c|}{0.804} & \multicolumn{1}{c|}{0.842} & \multicolumn{1}{c|}{0.794}  & 0.890 \\ \cline{2-14} 
                              & Deceptive Auxiliary Annotations              & \multicolumn{1}{c|}{0.090\%} & \multicolumn{1}{c|}{0.022\%}  & \multicolumn{1}{c|}{0.158\%}  & \multicolumn{1}{c|}{0.026} & \multicolumn{1}{c|}{0.015} & 0.037 & \multicolumn{1}{c|}{0.562} & \multicolumn{1}{c|}{0.386}  & \multicolumn{1}{c|}{0.738} & \multicolumn{1}{c|}{0.693} & \multicolumn{1}{c|}{0.557}  & 0.829 \\ \cline{2-14} 
                              & Modifying Legend                             & \multicolumn{1}{c|}{0.064\%} & \multicolumn{1}{c|}{0.012\%}  & \multicolumn{1}{c|}{0.116\%}  & \multicolumn{1}{c|}{0.023} & \multicolumn{1}{c|}{0.014} & 0.031 & \multicolumn{1}{c|}{0.763} & \multicolumn{1}{c|}{0.657}  & \multicolumn{1}{c|}{0.868} & \multicolumn{1}{c|}{0.858} & \multicolumn{1}{c|}{0.788}  & 0.929 \\ \cline{2-14} 
                              & Hiding Labels                                & \multicolumn{1}{c|}{0.083\%} & \multicolumn{1}{c|}{0.013\%}  & \multicolumn{1}{c|}{0.153\%}  & \multicolumn{1}{c|}{0.024} & \multicolumn{1}{c|}{0.012} & 0.036 & \multicolumn{1}{c|}{0.779} & \multicolumn{1}{c|}{0.731}  & \multicolumn{1}{c|}{0.827} & \multicolumn{1}{c|}{0.874} & \multicolumn{1}{c|}{0.843}  & 0.906 \\ \cline{2-14} 
                              & Adding or Removing Logos                     & \multicolumn{1}{c|}{0.021\%} & \multicolumn{1}{c|}{0.005\%}  & \multicolumn{1}{c|}{0.037\%}  & \multicolumn{1}{c|}{0.012} & \multicolumn{1}{c|}{0.006} & 0.018 & \multicolumn{1}{c|}{0.906} & \multicolumn{1}{c|}{0.875}  & \multicolumn{1}{c|}{0.938} & \multicolumn{1}{c|}{0.950} & \multicolumn{1}{c|}{0.932}  & 0.968 \\ \cline{2-14} 
                              & Data-Visual Disproportion                    & \multicolumn{1}{c|}{0.059\%} & \multicolumn{1}{c|}{0.045\%}  & \multicolumn{1}{c|}{0.073\%}  & \multicolumn{1}{c|}{0.024} & \multicolumn{1}{c|}{0.022} & 0.027 & \multicolumn{1}{c|}{0.750} & \multicolumn{1}{c|}{0.679}  & \multicolumn{1}{c|}{0.820} & \multicolumn{1}{c|}{0.853} & \multicolumn{1}{c|}{0.805}  & 0.901 \\ \cline{2-14} 
                              & Modifying Colormap                           & \multicolumn{1}{c|}{0.053\%} & \multicolumn{1}{c|}{0.000\%} & \multicolumn{1}{c|}{0.136\%}  & \multicolumn{1}{c|}{0.017} & \multicolumn{1}{c|}{0.005} & 0.029 & \multicolumn{1}{c|}{0.616} & \multicolumn{1}{c|}{0.473}  & \multicolumn{1}{c|}{0.759} & \multicolumn{1}{c|}{0.745} & \multicolumn{1}{c|}{0.635}  & 0.856 \\ \cline{2-14} 
                              & Mixture                                      & \multicolumn{1}{c|}{0.085\%} & \multicolumn{1}{c|}{0.016\%}  & \multicolumn{1}{c|}{0.154\%}  & \multicolumn{1}{c|}{0.025} & \multicolumn{1}{c|}{0.013} & 0.036 & \multicolumn{1}{c|}{0.802} & \multicolumn{1}{c|}{0.729}  & \multicolumn{1}{c|}{0.876} & \multicolumn{1}{c|}{0.887} & \multicolumn{1}{c|}{0.839}  & 0.935 \\ \hline
\multirow{10}{*}{EditGuard}   & Modifying Data Point Values                  & \multicolumn{1}{c|}{0.551\%} & \multicolumn{1}{c|}{0.398\%}  & \multicolumn{1}{c|}{0.704\%}  & \multicolumn{1}{c|}{0.073} & \multicolumn{1}{c|}{0.063} & 0.083 & \multicolumn{1}{c|}{0.343} & \multicolumn{1}{c|}{0.213}  & \multicolumn{1}{c|}{0.473} & \multicolumn{1}{c|}{0.486} & \multicolumn{1}{c|}{0.341}  & 0.631 \\ \cline{2-14} 
                              & Adding or Removing Data Points               & \multicolumn{1}{c|}{0.316\%} & \multicolumn{1}{c|}{0.075\%}  & \multicolumn{1}{c|}{0.557\%}  & \multicolumn{1}{c|}{0.048} & \multicolumn{1}{c|}{0.026} & 0.070 & \multicolumn{1}{c|}{0.733} & \multicolumn{1}{c|}{0.506}  & \multicolumn{1}{c|}{0.959} & \multicolumn{1}{c|}{0.799} & \multicolumn{1}{c|}{0.595}  & 1.003 \\ \cline{2-14} 
                              & Modifying Coordinate Values                  & \multicolumn{1}{c|}{0.534\%} & \multicolumn{1}{c|}{0.446\%}  & \multicolumn{1}{c|}{0.622\%}  & \multicolumn{1}{c|}{0.073} & \multicolumn{1}{c|}{0.067} & 0.079 & \multicolumn{1}{c|}{0.481} & \multicolumn{1}{c|}{0.365}  & \multicolumn{1}{c|}{0.597} & \multicolumn{1}{c|}{0.635} & \multicolumn{1}{c|}{0.530}  & 0.740 \\ \cline{2-14} 
                              & Deceptive Auxiliary Annotations              & \multicolumn{1}{c|}{0.476\%} & \multicolumn{1}{c|}{0.281\%}  & \multicolumn{1}{c|}{0.671\%}  & \multicolumn{1}{c|}{0.064} & \multicolumn{1}{c|}{0.045} & 0.083 & \multicolumn{1}{c|}{0.462} & \multicolumn{1}{c|}{0.260}  & \multicolumn{1}{c|}{0.664} & \multicolumn{1}{c|}{0.587} & \multicolumn{1}{c|}{0.400}  & 0.775 \\ \cline{2-14} 
                              & Modifying Legend                             & \multicolumn{1}{c|}{0.496\%} & \multicolumn{1}{c|}{0.292\%}  & \multicolumn{1}{c|}{0.700\%}  & \multicolumn{1}{c|}{0.065} & \multicolumn{1}{c|}{0.045} & 0.085 & \multicolumn{1}{c|}{0.546} & \multicolumn{1}{c|}{0.367}  & \multicolumn{1}{c|}{0.726} & \multicolumn{1}{c|}{0.675} & \multicolumn{1}{c|}{0.519}  & 0.831 \\ \cline{2-14} 
                              & Hiding Labels                                & \multicolumn{1}{c|}{0.316\%} & \multicolumn{1}{c|}{0.112\%}  & \multicolumn{1}{c|}{0.520\%}  & \multicolumn{1}{c|}{0.048} & \multicolumn{1}{c|}{0.027} & 0.070 & \multicolumn{1}{c|}{0.652} & \multicolumn{1}{c|}{0.510}  & \multicolumn{1}{c|}{0.794} & \multicolumn{1}{c|}{0.772} & \multicolumn{1}{c|}{0.655}  & 0.889 \\ \cline{2-14} 
                              & Adding or Removing Logos                     & \multicolumn{1}{c|}{0.369\%} & \multicolumn{1}{c|}{0.149\%}  & \multicolumn{1}{c|}{0.589\%}  & \multicolumn{1}{c|}{0.054} & \multicolumn{1}{c|}{0.032} & 0.075 & \multicolumn{1}{c|}{0.694} & \multicolumn{1}{c|}{0.524}  & \multicolumn{1}{c|}{0.863} & \multicolumn{1}{c|}{0.797} & \multicolumn{1}{c|}{0.671}  & 0.923 \\ \cline{2-14} 
                              & Data-Visual Disproportion                    & \multicolumn{1}{c|}{0.646\%} & \multicolumn{1}{c|}{0.512\%}  & \multicolumn{1}{c|}{0.780\%}  & \multicolumn{1}{c|}{0.080} & \multicolumn{1}{c|}{0.072} & 0.088 & \multicolumn{1}{c|}{0.444} & \multicolumn{1}{c|}{0.320}  & \multicolumn{1}{c|}{0.567} & \multicolumn{1}{c|}{0.596} & \multicolumn{1}{c|}{0.470}  & 0.722 \\ \cline{2-14} 
                              & Modifying Colormap                           & \multicolumn{1}{c|}{0.494\%} & \multicolumn{1}{c|}{0.250\%}  & \multicolumn{1}{c|}{0.738\%}  & \multicolumn{1}{c|}{0.064} & \multicolumn{1}{c|}{0.043} & 0.086 & \multicolumn{1}{c|}{0.407} & \multicolumn{1}{c|}{0.187}  & \multicolumn{1}{c|}{0.627} & \multicolumn{1}{c|}{0.521} & \multicolumn{1}{c|}{0.309}  & 0.733 \\ \cline{2-14} 
                              & Mixture                                      & \multicolumn{1}{c|}{0.401\%} & \multicolumn{1}{c|}{0.117\%}  & \multicolumn{1}{c|}{0.685\%}  & \multicolumn{1}{c|}{0.055} & \multicolumn{1}{c|}{0.032} & 0.079 & \multicolumn{1}{c|}{0.693} & \multicolumn{1}{c|}{0.544}  & \multicolumn{1}{c|}{0.842} & \multicolumn{1}{c|}{0.801} & \multicolumn{1}{c|}{0.684}  & 0.918 \\ \hline
\multirow{10}{*}{ManTraNet}   & Modifying Data Point Values                  & \multicolumn{1}{c|}{1.876\%} & \multicolumn{1}{c|}{1.373\%}  & \multicolumn{1}{c|}{2.378\%}  & \multicolumn{1}{c|}{0.135} & \multicolumn{1}{c|}{0.117} & 0.153 & \multicolumn{1}{c|}{0.068} & \multicolumn{1}{c|}{0.025}  & \multicolumn{1}{c|}{0.112} & \multicolumn{1}{c|}{0.122} & \multicolumn{1}{c|}{0.047}  & 0.197 \\ \cline{2-14} 
                              & Adding or Removing Data Points               & \multicolumn{1}{c|}{2.389\%} & \multicolumn{1}{c|}{1.821\%}  & \multicolumn{1}{c|}{2.958\%}  & \multicolumn{1}{c|}{0.153} & \multicolumn{1}{c|}{0.134} & 0.171 & \multicolumn{1}{c|}{0.043} & \multicolumn{1}{c|}{0.000} & \multicolumn{1}{c|}{0.107} & \multicolumn{1}{c|}{0.073} & \multicolumn{1}{c|}{0.000} & 0.172 \\ \cline{2-14} 
                              & Modifying Coordinate Values                  & \multicolumn{1}{c|}{1.887\%} & \multicolumn{1}{c|}{1.096\%}  & \multicolumn{1}{c|}{2.677\%}  & \multicolumn{1}{c|}{0.133} & \multicolumn{1}{c|}{0.107} & 0.159 & \multicolumn{1}{c|}{0.001} & \multicolumn{1}{c|}{0.000} & \multicolumn{1}{c|}{0.002} & \multicolumn{1}{c|}{0.002} & \multicolumn{1}{c|}{0.000} & 0.004 \\ \cline{2-14} 
                              & Deceptive Auxiliary Annotations              & \multicolumn{1}{c|}{1.508\%} & \multicolumn{1}{c|}{0.821\%}  & \multicolumn{1}{c|}{2.195\%}  & \multicolumn{1}{c|}{0.117} & \multicolumn{1}{c|}{0.088} & 0.146 & \multicolumn{1}{c|}{0.180} & \multicolumn{1}{c|}{0.040}  & \multicolumn{1}{c|}{0.321} & \multicolumn{1}{c|}{0.269} & \multicolumn{1}{c|}{0.094}  & 0.444 \\ \cline{2-14} 
                              & Modifying Legend                             & \multicolumn{1}{c|}{2.716\%} & \multicolumn{1}{c|}{1.180\%}  & \multicolumn{1}{c|}{4.253\%}  & \multicolumn{1}{c|}{0.157} & \multicolumn{1}{c|}{0.120} & 0.195 & \multicolumn{1}{c|}{0.008} & \multicolumn{1}{c|}{0.000} & \multicolumn{1}{c|}{0.024} & \multicolumn{1}{c|}{0.016} & \multicolumn{1}{c|}{0.000} & 0.044 \\ \cline{2-14} 
                              & Hiding Labels                                & \multicolumn{1}{c|}{1.627\%} & \multicolumn{1}{c|}{1.031\%}  & \multicolumn{1}{c|}{2.223\%}  & \multicolumn{1}{c|}{0.123} & \multicolumn{1}{c|}{0.098} & 0.148 & \multicolumn{1}{c|}{0.018} & \multicolumn{1}{c|}{0.000} & \multicolumn{1}{c|}{0.038} & \multicolumn{1}{c|}{0.035} & \multicolumn{1}{c|}{0.000} & 0.072 \\ \cline{2-14} 
                              & Adding or Removing Logos                     & \multicolumn{1}{c|}{2.563\%} & \multicolumn{1}{c|}{1.297\%}  & \multicolumn{1}{c|}{3.830\%}  & \multicolumn{1}{c|}{0.152} & \multicolumn{1}{c|}{0.113} & 0.190 & \multicolumn{1}{c|}{0.069} & \multicolumn{1}{c|}{0.008}  & \multicolumn{1}{c|}{0.131} & \multicolumn{1}{c|}{0.120} & \multicolumn{1}{c|}{0.020}  & 0.219 \\ \cline{2-14} 
                              & Data-Visual Disproportion                    & \multicolumn{1}{c|}{2.384\%} & \multicolumn{1}{c|}{2.010\%}  & \multicolumn{1}{c|}{2.759\%}  & \multicolumn{1}{c|}{0.154} & \multicolumn{1}{c|}{0.142} & 0.166 & \multicolumn{1}{c|}{0.015} & \multicolumn{1}{c|}{0.000} & \multicolumn{1}{c|}{0.037} & \multicolumn{1}{c|}{0.028} & \multicolumn{1}{c|}{0.000} & 0.068 \\ \cline{2-14} 
                              & Modifying Colormap                           & \multicolumn{1}{c|}{3.416\%} & \multicolumn{1}{c|}{1.282\%}  & \multicolumn{1}{c|}{5.550\%}  & \multicolumn{1}{c|}{0.169} & \multicolumn{1}{c|}{0.112} & 0.226 & \multicolumn{1}{c|}{0.094} & \multicolumn{1}{c|}{0.000} & \multicolumn{1}{c|}{0.225} & \multicolumn{1}{c|}{0.135} & \multicolumn{1}{c|}{0.000} & 0.312 \\ \cline{2-14} 
                              & Mixture                                      & \multicolumn{1}{c|}{3.713\%} & \multicolumn{1}{c|}{0.754\%}  & \multicolumn{1}{c|}{6.671\%}  & \multicolumn{1}{c|}{0.174} & \multicolumn{1}{c|}{0.112} & 0.236 & \multicolumn{1}{c|}{0.041} & \multicolumn{1}{c|}{0.000} & \multicolumn{1}{c|}{0.096} & \multicolumn{1}{c|}{0.071} & \multicolumn{1}{c|}{0.000} & 0.160 \\ \hline
\multirow{10}{*}{MVSS-Net}    & Modifying Data Point Values                  & \multicolumn{1}{c|}{3.220\%} & \multicolumn{1}{c|}{0.882\%}  & \multicolumn{1}{c|}{5.558\%}  & \multicolumn{1}{c|}{0.154} & \multicolumn{1}{c|}{0.086} & 0.223 & \multicolumn{1}{c|}{0.010} & \multicolumn{1}{c|}{0.002}  & \multicolumn{1}{c|}{0.018} & \multicolumn{1}{c|}{0.031} & \multicolumn{1}{c|}{0.000} & 0.086 \\ \cline{2-14} 
                              & Adding or Removing Data Points               & \multicolumn{1}{c|}{3.821\%} & \multicolumn{1}{c|}{0.953\%}  & \multicolumn{1}{c|}{6.688\%}  & \multicolumn{1}{c|}{0.160} & \multicolumn{1}{c|}{0.076} & 0.245 & \multicolumn{1}{c|}{0.033} & \multicolumn{1}{c|}{0.011}  & \multicolumn{1}{c|}{0.055} & \multicolumn{1}{c|}{0.076} & \multicolumn{1}{c|}{0.000} & 0.177 \\ \cline{2-14} 
                              & Modifying Coordinate Values                  & \multicolumn{1}{c|}{4.891\%} & \multicolumn{1}{c|}{1.430\%}  & \multicolumn{1}{c|}{8.352\%}  & \multicolumn{1}{c|}{0.190} & \multicolumn{1}{c|}{0.106} & 0.275 & \multicolumn{1}{c|}{0.010} & \multicolumn{1}{c|}{0.004}  & \multicolumn{1}{c|}{0.017} & \multicolumn{1}{c|}{0.003} & \multicolumn{1}{c|}{0.000} & 0.008 \\ \cline{2-14} 
                              & Deceptive Auxiliary Annotations              & \multicolumn{1}{c|}{3.691\%} & \multicolumn{1}{c|}{0.873\%}  & \multicolumn{1}{c|}{6.509\%}  & \multicolumn{1}{c|}{0.154} & \multicolumn{1}{c|}{0.068} & 0.241 & \multicolumn{1}{c|}{0.034} & \multicolumn{1}{c|}{0.000} & \multicolumn{1}{c|}{0.072} & \multicolumn{1}{c|}{0.098} & \multicolumn{1}{c|}{0.000} & 0.226 \\ \cline{2-14} 
                              & Modifying Legend                             & \multicolumn{1}{c|}{2.585\%} & \multicolumn{1}{c|}{0.322\%}  & \multicolumn{1}{c|}{4.848\%}  & \multicolumn{1}{c|}{0.130} & \multicolumn{1}{c|}{0.059} & 0.201 & \multicolumn{1}{c|}{0.006} & \multicolumn{1}{c|}{0.002}  & \multicolumn{1}{c|}{0.011} & \multicolumn{1}{c|}{0.006} & \multicolumn{1}{c|}{0.000} & 0.015 \\ \cline{2-14} 
                              & Hiding Labels                                & \multicolumn{1}{c|}{3.555\%} & \multicolumn{1}{c|}{0.000\%} & \multicolumn{1}{c|}{7.381\%}  & \multicolumn{1}{c|}{0.156} & \multicolumn{1}{c|}{0.075} & 0.236 & \multicolumn{1}{c|}{0.015} & \multicolumn{1}{c|}{0.008}  & \multicolumn{1}{c|}{0.022} & \multicolumn{1}{c|}{0.077} & \multicolumn{1}{c|}{0.000} & 0.162 \\ \cline{2-14} 
                              & Adding or Removing Logos                     & \multicolumn{1}{c|}{4.204\%} & \multicolumn{1}{c|}{1.687\%}  & \multicolumn{1}{c|}{6.720\%}  & \multicolumn{1}{c|}{0.184} & \multicolumn{1}{c|}{0.115} & 0.252 & \multicolumn{1}{c|}{0.012} & \multicolumn{1}{c|}{0.008}  & \multicolumn{1}{c|}{0.016} & \multicolumn{1}{c|}{0.100} & \multicolumn{1}{c|}{0.000} & 0.230 \\ \cline{2-14} 
                              & Data-Visual Disproportion                    & \multicolumn{1}{c|}{3.411\%} & \multicolumn{1}{c|}{1.413\%}  & \multicolumn{1}{c|}{5.410\%}  & \multicolumn{1}{c|}{0.168} & \multicolumn{1}{c|}{0.110} & 0.226 & \multicolumn{1}{c|}{0.010} & \multicolumn{1}{c|}{0.003}  & \multicolumn{1}{c|}{0.016} & \multicolumn{1}{c|}{0.005} & \multicolumn{1}{c|}{0.000} & 0.018 \\ \cline{2-14} 
                              & Modifying Colormap                           & \multicolumn{1}{c|}{8.136\%} & \multicolumn{1}{c|}{2.293\%}  & \multicolumn{1}{c|}{13.978\%} & \multicolumn{1}{c|}{0.237} & \multicolumn{1}{c|}{0.118} & 0.357 & \multicolumn{1}{c|}{0.031} & \multicolumn{1}{c|}{0.000} & \multicolumn{1}{c|}{0.072} & \multicolumn{1}{c|}{0.014} & \multicolumn{1}{c|}{0.000} & 0.045 \\ \cline{2-14} 
                              & Mixture                                      & \multicolumn{1}{c|}{3.859\%} & \multicolumn{1}{c|}{1.139\%}  & \multicolumn{1}{c|}{6.579\%}  & \multicolumn{1}{c|}{0.169} & \multicolumn{1}{c|}{0.093} & 0.245 & \multicolumn{1}{c|}{0.022} & \multicolumn{1}{c|}{0.011}  & \multicolumn{1}{c|}{0.033} & \multicolumn{1}{c|}{0.051} & \multicolumn{1}{c|}{0.000} & 0.142 \\ \hline
\end{tabular}
\end{table}

\vfill
\newpage

\begin{table}[!h]
\renewcommand{\arraystretch}{1.5}
\centering
\caption{\label{supp_tab_3}The accuracy evaluation results of tampering detection on the MCD. The Mean column represents the average value of each metric (Noise Percentage, RMSE, IoU, F1 Score). The Lower and Upper columns indicate the 95\% confidence interval for each metric.}

\fontsize{6.5}{7.8}\selectfont
\begin{tabular}{|c|l|cccccc|cccccc|}
\hline
\multirow{3}{*}{Method}      & \multicolumn{1}{c|}{\multirow{3}{*}{Subset}} & \multicolumn{6}{c|}{No Tampering}                                                                                                                            & \multicolumn{6}{c|}{Post-Tampering}                                                                                                                     \\ \cline{3-14} 
                             & \multicolumn{1}{c|}{}                        & \multicolumn{3}{c|}{Noise Percentage $\downarrow$}                                                      & \multicolumn{3}{c|}{RMSE $\downarrow$}                                       & \multicolumn{3}{c|}{IoU $\uparrow$}                                                             & \multicolumn{3}{c|}{F1 Score $\uparrow$}                                    \\ \cline{3-14} 
                             & \multicolumn{1}{c|}{}                        & \multicolumn{1}{c|}{Mean}    & \multicolumn{1}{c|}{Lower}   & \multicolumn{1}{c|}{Upper}   & \multicolumn{1}{c|}{Mean}  & \multicolumn{1}{c|}{Lower} & Upper & \multicolumn{1}{c|}{Mean}  & \multicolumn{1}{c|}{Lower} & \multicolumn{1}{c|}{Upper} & \multicolumn{1}{c|}{Mean}  & \multicolumn{1}{c|}{Lower}  & Upper \\ \hline
\multirow{9}{*}{\textbf{VizDefender}} & Textual Tampering - 1                        & \multicolumn{1}{c|}{0.036\%} & \multicolumn{1}{c|}{0.022\%} & \multicolumn{1}{c|}{0.049\%} & \multicolumn{1}{c|}{0.013} & \multicolumn{1}{c|}{0.010} & 0.016 & \multicolumn{1}{c|}{0.549} & \multicolumn{1}{c|}{0.494} & \multicolumn{1}{c|}{0.604} & \multicolumn{1}{c|}{0.660} & \multicolumn{1}{c|}{0.604}  & 0.716 \\ \cline{2-14} 
                             & Textual Tampering - 2                        & \multicolumn{1}{c|}{0.035\%} & \multicolumn{1}{c|}{0.021\%} & \multicolumn{1}{c|}{0.048\%} & \multicolumn{1}{c|}{0.013} & \multicolumn{1}{c|}{0.010} & 0.015 & \multicolumn{1}{c|}{0.598} & \multicolumn{1}{c|}{0.553} & \multicolumn{1}{c|}{0.643} & \multicolumn{1}{c|}{0.719} & \multicolumn{1}{c|}{0.677}  & 0.761 \\ \cline{2-14} 
                             & Textual Tampering - 3                        & \multicolumn{1}{c|}{0.035\%} & \multicolumn{1}{c|}{0.021\%} & \multicolumn{1}{c|}{0.048\%} & \multicolumn{1}{c|}{0.013} & \multicolumn{1}{c|}{0.010} & 0.015 & \multicolumn{1}{c|}{0.624} & \multicolumn{1}{c|}{0.583} & \multicolumn{1}{c|}{0.665} & \multicolumn{1}{c|}{0.746} & \multicolumn{1}{c|}{0.709}  & 0.782 \\ \cline{2-14} 
                             & Graphical Tampering - 1                      & \multicolumn{1}{c|}{0.034\%} & \multicolumn{1}{c|}{0.017\%} & \multicolumn{1}{c|}{0.051\%} & \multicolumn{1}{c|}{0.011} & \multicolumn{1}{c|}{0.008} & 0.014 & \multicolumn{1}{c|}{0.552} & \multicolumn{1}{c|}{0.498} & \multicolumn{1}{c|}{0.606} & \multicolumn{1}{c|}{0.665} & \multicolumn{1}{c|}{0.612}  & 0.718 \\ \cline{2-14} 
                             & Graphical Tampering - 2                      & \multicolumn{1}{c|}{0.034\%} & \multicolumn{1}{c|}{0.017\%} & \multicolumn{1}{c|}{0.051\%} & \multicolumn{1}{c|}{0.011} & \multicolumn{1}{c|}{0.008} & 0.014 & \multicolumn{1}{c|}{0.595} & \multicolumn{1}{c|}{0.551} & \multicolumn{1}{c|}{0.639} & \multicolumn{1}{c|}{0.718} & \multicolumn{1}{c|}{0.678}  & 0.759 \\ \cline{2-14} 
                             & Graphical Tampering - 3                      & \multicolumn{1}{c|}{0.036\%} & \multicolumn{1}{c|}{0.018\%} & \multicolumn{1}{c|}{0.053\%} & \multicolumn{1}{c|}{0.012} & \multicolumn{1}{c|}{0.009} & 0.015 & \multicolumn{1}{c|}{0.606} & \multicolumn{1}{c|}{0.563} & \multicolumn{1}{c|}{0.649} & \multicolumn{1}{c|}{0.729} & \multicolumn{1}{c|}{0.690}  & 0.767 \\ \cline{2-14} 
                             & Painting - 1                                 & \multicolumn{1}{c|}{0.049\%} & \multicolumn{1}{c|}{0.025\%} & \multicolumn{1}{c|}{0.073\%} & \multicolumn{1}{c|}{0.014} & \multicolumn{1}{c|}{0.011} & 0.018 & \multicolumn{1}{c|}{0.290} & \multicolumn{1}{c|}{0.261} & \multicolumn{1}{c|}{0.319} & \multicolumn{1}{c|}{0.430} & \multicolumn{1}{c|}{0.395}  & 0.465 \\ \cline{2-14} 
                             & Painting - 2                                 & \multicolumn{1}{c|}{0.049\%} & \multicolumn{1}{c|}{0.025\%} & \multicolumn{1}{c|}{0.073\%} & \multicolumn{1}{c|}{0.014} & \multicolumn{1}{c|}{0.011} & 0.018 & \multicolumn{1}{c|}{0.294} & \multicolumn{1}{c|}{0.269} & \multicolumn{1}{c|}{0.319} & \multicolumn{1}{c|}{0.439} & \multicolumn{1}{c|}{0.409}  & 0.470 \\ \cline{2-14} 
                             & Painting - 3                                 & \multicolumn{1}{c|}{0.049\%} & \multicolumn{1}{c|}{0.025\%} & \multicolumn{1}{c|}{0.073\%} & \multicolumn{1}{c|}{0.014} & \multicolumn{1}{c|}{0.011} & 0.018 & \multicolumn{1}{c|}{0.303} & \multicolumn{1}{c|}{0.279} & \multicolumn{1}{c|}{0.326} & \multicolumn{1}{c|}{0.452} & \multicolumn{1}{c|}{0.424}  & 0.480 \\ \hline
\multirow{9}{*}{EditGuard}   & Textual Tampering - 1                        & \multicolumn{1}{c|}{0.635\%} & \multicolumn{1}{c|}{0.574\%} & \multicolumn{1}{c|}{0.697\%} & \multicolumn{1}{c|}{0.076} & \multicolumn{1}{c|}{0.072} & 0.081 & \multicolumn{1}{c|}{0.120} & \multicolumn{1}{c|}{0.093} & \multicolumn{1}{c|}{0.147} & \multicolumn{1}{c|}{0.193} & \multicolumn{1}{c|}{0.157}  & 0.228 \\ \cline{2-14} 
                             & Textual Tampering - 2                        & \multicolumn{1}{c|}{0.622\%} & \multicolumn{1}{c|}{0.561\%} & \multicolumn{1}{c|}{0.684\%} & \multicolumn{1}{c|}{0.075} & \multicolumn{1}{c|}{0.071} & 0.080 & \multicolumn{1}{c|}{0.195} & \multicolumn{1}{c|}{0.159} & \multicolumn{1}{c|}{0.231} & \multicolumn{1}{c|}{0.295} & \multicolumn{1}{c|}{0.253}  & 0.337 \\ \cline{2-14} 
                             & Textual Tampering - 3                        & \multicolumn{1}{c|}{0.625\%} & \multicolumn{1}{c|}{0.563\%} & \multicolumn{1}{c|}{0.687\%} & \multicolumn{1}{c|}{0.076} & \multicolumn{1}{c|}{0.071} & 0.080 & \multicolumn{1}{c|}{0.246} & \multicolumn{1}{c|}{0.209} & \multicolumn{1}{c|}{0.282} & \multicolumn{1}{c|}{0.363} & \multicolumn{1}{c|}{0.320}  & 0.406 \\ \cline{2-14} 
                             & Graphical Tampering - 1                      & \multicolumn{1}{c|}{0.577\%} & \multicolumn{1}{c|}{0.517\%} & \multicolumn{1}{c|}{0.637\%} & \multicolumn{1}{c|}{0.072} & \multicolumn{1}{c|}{0.067} & 0.077 & \multicolumn{1}{c|}{0.294} & \multicolumn{1}{c|}{0.248} & \multicolumn{1}{c|}{0.340} & \multicolumn{1}{c|}{0.407} & \multicolumn{1}{c|}{0.352}  & 0.461 \\ \cline{2-14} 
                             & Graphical Tampering - 2                      & \multicolumn{1}{c|}{0.577\%} & \multicolumn{1}{c|}{0.517\%} & \multicolumn{1}{c|}{0.637\%} & \multicolumn{1}{c|}{0.072} & \multicolumn{1}{c|}{0.067} & 0.077 & \multicolumn{1}{c|}{0.388} & \multicolumn{1}{c|}{0.340} & \multicolumn{1}{c|}{0.435} & \multicolumn{1}{c|}{0.513} & \multicolumn{1}{c|}{0.460}  & 0.567 \\ \cline{2-14} 
                             & Graphical Tampering - 3                      & \multicolumn{1}{c|}{0.585\%} & \multicolumn{1}{c|}{0.524\%} & \multicolumn{1}{c|}{0.647\%} & \multicolumn{1}{c|}{0.073} & \multicolumn{1}{c|}{0.068} & 0.077 & \multicolumn{1}{c|}{0.431} & \multicolumn{1}{c|}{0.382} & \multicolumn{1}{c|}{0.479} & \multicolumn{1}{c|}{0.557} & \multicolumn{1}{c|}{0.504}  & 0.610 \\ \cline{2-14} 
                             & Painting - 1                                 & \multicolumn{1}{c|}{0.614\%} & \multicolumn{1}{c|}{0.553\%} & \multicolumn{1}{c|}{0.675\%} & \multicolumn{1}{c|}{0.075} & \multicolumn{1}{c|}{0.070} & 0.079 & \multicolumn{1}{c|}{0.117} & \multicolumn{1}{c|}{0.095} & \multicolumn{1}{c|}{0.138} & \multicolumn{1}{c|}{0.194} & \multicolumn{1}{c|}{0.163}  & 0.225 \\ \cline{2-14} 
                             & Painting - 2                                 & \multicolumn{1}{c|}{0.617\%} & \multicolumn{1}{c|}{0.555\%} & \multicolumn{1}{c|}{0.678\%} & \multicolumn{1}{c|}{0.075} & \multicolumn{1}{c|}{0.070} & 0.080 & \multicolumn{1}{c|}{0.147} & \multicolumn{1}{c|}{0.125} & \multicolumn{1}{c|}{0.169} & \multicolumn{1}{c|}{0.241} & \multicolumn{1}{c|}{0.210}  & 0.272 \\ \cline{2-14} 
                             & Painting - 3                                 & \multicolumn{1}{c|}{0.616\%} & \multicolumn{1}{c|}{0.554\%} & \multicolumn{1}{c|}{0.677\%} & \multicolumn{1}{c|}{0.075} & \multicolumn{1}{c|}{0.070} & 0.079 & \multicolumn{1}{c|}{0.177} & \multicolumn{1}{c|}{0.156} & \multicolumn{1}{c|}{0.199} & \multicolumn{1}{c|}{0.288} & \multicolumn{1}{c|}{0.258}  & 0.317 \\ \hline
\multirow{9}{*}{ManTraNet}   & Textual Tampering - 1                        & \multicolumn{1}{c|}{4.256\%} & \multicolumn{1}{c|}{3.537\%} & \multicolumn{1}{c|}{4.976\%} & \multicolumn{1}{c|}{0.191} & \multicolumn{1}{c|}{0.176} & 0.207 & \multicolumn{1}{c|}{0.001} & \multicolumn{1}{c|}{0.000} & \multicolumn{1}{c|}{0.001} & \multicolumn{1}{c|}{0.001} & \multicolumn{1}{c|}{0.000}  & 0.002 \\ \cline{2-14} 
                             & Textual Tampering - 2                        & \multicolumn{1}{c|}{4.194\%} & \multicolumn{1}{c|}{3.474\%} & \multicolumn{1}{c|}{4.915\%} & \multicolumn{1}{c|}{0.190} & \multicolumn{1}{c|}{0.174} & 0.205 & \multicolumn{1}{c|}{0.001} & \multicolumn{1}{c|}{0.000} & \multicolumn{1}{c|}{0.002} & \multicolumn{1}{c|}{0.003} & \multicolumn{1}{c|}{0.000}  & 0.005 \\ \cline{2-14} 
                             & Textual Tampering - 3                        & \multicolumn{1}{c|}{4.225\%} & \multicolumn{1}{c|}{3.508\%} & \multicolumn{1}{c|}{4.943\%} & \multicolumn{1}{c|}{0.191} & \multicolumn{1}{c|}{0.175} & 0.206 & \multicolumn{1}{c|}{0.003} & \multicolumn{1}{c|}{0.001} & \multicolumn{1}{c|}{0.006} & \multicolumn{1}{c|}{0.006} & \multicolumn{1}{c|}{0.001}  & 0.010 \\ \cline{2-14} 
                             & Graphical Tampering - 1                      & \multicolumn{1}{c|}{4.168\%} & \multicolumn{1}{c|}{3.519\%} & \multicolumn{1}{c|}{4.817\%} & \multicolumn{1}{c|}{0.190} & \multicolumn{1}{c|}{0.176} & 0.205 & \multicolumn{1}{c|}{0.017} & \multicolumn{1}{c|}{0.012} & \multicolumn{1}{c|}{0.022} & \multicolumn{1}{c|}{0.033} & \multicolumn{1}{c|}{0.024}  & 0.042 \\ \cline{2-14} 
                             & Graphical Tampering - 2                      & \multicolumn{1}{c|}{4.168\%} & \multicolumn{1}{c|}{3.519\%} & \multicolumn{1}{c|}{4.817\%} & \multicolumn{1}{c|}{0.190} & \multicolumn{1}{c|}{0.176} & 0.205 & \multicolumn{1}{c|}{0.043} & \multicolumn{1}{c|}{0.032} & \multicolumn{1}{c|}{0.054} & \multicolumn{1}{c|}{0.077} & \multicolumn{1}{c|}{0.059}  & 0.095 \\ \cline{2-14} 
                             & Graphical Tampering - 3                      & \multicolumn{1}{c|}{4.170\%} & \multicolumn{1}{c|}{3.521\%} & \multicolumn{1}{c|}{4.819\%} & \multicolumn{1}{c|}{0.190} & \multicolumn{1}{c|}{0.176} & 0.205 & \multicolumn{1}{c|}{0.056} & \multicolumn{1}{c|}{0.042} & \multicolumn{1}{c|}{0.070} & \multicolumn{1}{c|}{0.099} & \multicolumn{1}{c|}{0.078}  & 0.120 \\ \cline{2-14} 
                             & Painting - 1                                 & \multicolumn{1}{c|}{5.140\%} & \multicolumn{1}{c|}{4.155\%} & \multicolumn{1}{c|}{6.125\%} & \multicolumn{1}{c|}{0.206} & \multicolumn{1}{c|}{0.188} & 0.225 & \multicolumn{1}{c|}{0.042} & \multicolumn{1}{c|}{0.030} & \multicolumn{1}{c|}{0.054} & \multicolumn{1}{c|}{0.075} & \multicolumn{1}{c|}{0.056}  & 0.095 \\ \cline{2-14} 
                             & Painting - 2                                 & \multicolumn{1}{c|}{5.127\%} & \multicolumn{1}{c|}{4.144\%} & \multicolumn{1}{c|}{6.111\%} & \multicolumn{1}{c|}{0.206} & \multicolumn{1}{c|}{0.187} & 0.225 & \multicolumn{1}{c|}{0.072} & \multicolumn{1}{c|}{0.052} & \multicolumn{1}{c|}{0.092} & \multicolumn{1}{c|}{0.121} & \multicolumn{1}{c|}{0.092}  & 0.150 \\ \cline{2-14} 
                             & Painting - 3                                 & \multicolumn{1}{c|}{5.126\%} & \multicolumn{1}{c|}{4.142\%} & \multicolumn{1}{c|}{6.110\%} & \multicolumn{1}{c|}{0.206} & \multicolumn{1}{c|}{0.187} & 0.225 & \multicolumn{1}{c|}{0.097} & \multicolumn{1}{c|}{0.074} & \multicolumn{1}{c|}{0.119} & \multicolumn{1}{c|}{0.159} & \multicolumn{1}{c|}{0.126}  & 0.192 \\ \hline
\multirow{9}{*}{MVSS-Net}    & Textual Tampering - 1                        & \multicolumn{1}{c|}{4.275\%} & \multicolumn{1}{c|}{2.870\%} & \multicolumn{1}{c|}{5.681\%} & \multicolumn{1}{c|}{0.142} & \multicolumn{1}{c|}{0.111} & 0.172 & \multicolumn{1}{c|}{0.001} & \multicolumn{1}{c|}{0.001} & \multicolumn{1}{c|}{0.001} & \multicolumn{1}{c|}{0.001} & \multicolumn{1}{c|}{0.000} & 0.002 \\ \cline{2-14} 
                             & Textual Tampering - 2                        & \multicolumn{1}{c|}{4.206\%} & \multicolumn{1}{c|}{2.799\%} & \multicolumn{1}{c|}{5.614\%} & \multicolumn{1}{c|}{0.139} & \multicolumn{1}{c|}{0.109} & 0.169 & \multicolumn{1}{c|}{0.002} & \multicolumn{1}{c|}{0.001} & \multicolumn{1}{c|}{0.004} & \multicolumn{1}{c|}{0.002} & \multicolumn{1}{c|}{0.000}  & 0.004 \\ \cline{2-14} 
                             & Textual Tampering - 3                        & \multicolumn{1}{c|}{4.206\%} & \multicolumn{1}{c|}{2.799\%} & \multicolumn{1}{c|}{5.614\%} & \multicolumn{1}{c|}{0.139} & \multicolumn{1}{c|}{0.109} & 0.169 & \multicolumn{1}{c|}{0.003} & \multicolumn{1}{c|}{0.001} & \multicolumn{1}{c|}{0.005} & \multicolumn{1}{c|}{0.002} & \multicolumn{1}{c|}{0.000} & 0.005 \\ \cline{2-14} 
                             & Graphical Tampering - 1                      & \multicolumn{1}{c|}{4.448\%} & \multicolumn{1}{c|}{3.178\%} & \multicolumn{1}{c|}{5.718\%} & \multicolumn{1}{c|}{0.156} & \multicolumn{1}{c|}{0.128} & 0.184 & \multicolumn{1}{c|}{0.006} & \multicolumn{1}{c|}{0.004} & \multicolumn{1}{c|}{0.007} & \multicolumn{1}{c|}{0.064} & \multicolumn{1}{c|}{0.038}  & 0.091 \\ \cline{2-14} 
                             & Graphical Tampering - 2                      & \multicolumn{1}{c|}{4.448\%} & \multicolumn{1}{c|}{3.178\%} & \multicolumn{1}{c|}{5.718\%} & \multicolumn{1}{c|}{0.156} & \multicolumn{1}{c|}{0.128} & 0.184 & \multicolumn{1}{c|}{0.012} & \multicolumn{1}{c|}{0.009} & \multicolumn{1}{c|}{0.014} & \multicolumn{1}{c|}{0.121} & \multicolumn{1}{c|}{0.083}  & 0.160 \\ \cline{2-14} 
                             & Graphical Tampering - 3                      & \multicolumn{1}{c|}{4.520\%} & \multicolumn{1}{c|}{3.252\%} & \multicolumn{1}{c|}{5.789\%} & \multicolumn{1}{c|}{0.159} & \multicolumn{1}{c|}{0.130} & 0.187 & \multicolumn{1}{c|}{0.017} & \multicolumn{1}{c|}{0.014} & \multicolumn{1}{c|}{0.021} & \multicolumn{1}{c|}{0.164} & \multicolumn{1}{c|}{0.126}  & 0.202 \\ \cline{2-14} 
                             & Painting - 1                                 & \multicolumn{1}{c|}{4.378\%} & \multicolumn{1}{c|}{2.793\%} & \multicolumn{1}{c|}{5.963\%} & \multicolumn{1}{c|}{0.147} & \multicolumn{1}{c|}{0.117} & 0.177 & \multicolumn{1}{c|}{0.002} & \multicolumn{1}{c|}{0.001} & \multicolumn{1}{c|}{0.002} & \multicolumn{1}{c|}{0.070} & \multicolumn{1}{c|}{0.043}  & 0.096 \\ \cline{2-14} 
                             & Painting - 2                                 & \multicolumn{1}{c|}{4.438\%} & \multicolumn{1}{c|}{2.855\%} & \multicolumn{1}{c|}{6.021\%} & \multicolumn{1}{c|}{0.149} & \multicolumn{1}{c|}{0.120} & 0.179 & \multicolumn{1}{c|}{0.005} & \multicolumn{1}{c|}{0.002} & \multicolumn{1}{c|}{0.008} & \multicolumn{1}{c|}{0.074} & \multicolumn{1}{c|}{0.053}  & 0.094 \\ \cline{2-14} 
                             & Painting - 3                                 & \multicolumn{1}{c|}{4.438\%} & \multicolumn{1}{c|}{2.855\%} & \multicolumn{1}{c|}{6.021\%} & \multicolumn{1}{c|}{0.149} & \multicolumn{1}{c|}{0.120} & 0.179 & \multicolumn{1}{c|}{0.007} & \multicolumn{1}{c|}{0.005} & \multicolumn{1}{c|}{0.008} & \multicolumn{1}{c|}{0.084} & \multicolumn{1}{c|}{0.062}  & 0.106 \\ \hline
\end{tabular}
\end{table}

\vfill
\newpage

\subsection{Accuracy of Intent Analysis}

\begin{table}[!h]
\renewcommand{\arraystretch}{1.5}
\centering
\caption{\label{supp_tab_5}The accuracy evaluation results of tampering intent analysis on the MCD. The Mean column represents the average value of each metric (Cosine Similarity of Tampering Method and Tampering Intent). The Lower and Upper columns indicate the 95\% confidence interval for each metric.}

\small

\begin{tabular}{|c|l|llc|llc|}
\hline
\multirow{2}{*}{Method}                & \multicolumn{1}{c|}{\multirow{2}{*}{Tampering Type}} & \multicolumn{3}{c|}{Tamering Process   Similarity}                 & \multicolumn{3}{c|}{Tamering Intent   Similarity}                  \\ \cline{3-8} 
                                       & \multicolumn{1}{c|}{}                                & \multicolumn{1}{c|}{Mean}   & \multicolumn{1}{c|}{Lower}  & Upper  & \multicolumn{1}{c|}{Mean}   & \multicolumn{1}{c|}{Lower}  & Upper  \\ \hline
\multirow{10}{*}{\textbf{VizDefender}} & Modifying Data Point Values                          & \multicolumn{1}{l|}{0.8569} & \multicolumn{1}{l|}{0.8266} & 0.8873 & \multicolumn{1}{l|}{0.9021} & \multicolumn{1}{l|}{0.8749} & 0.9292 \\ \cline{2-8} 
                                       & Adding or Removing Data Points                       & \multicolumn{1}{l|}{0.8751} & \multicolumn{1}{l|}{0.8466} & 0.9037 & \multicolumn{1}{l|}{0.9000} & \multicolumn{1}{l|}{0.8850} & 0.9151 \\ \cline{2-8} 
                                       & Modifying Coordinate Values                          & \multicolumn{1}{l|}{0.8590} & \multicolumn{1}{l|}{0.8310} & 0.8871 & \multicolumn{1}{l|}{0.9133} & \multicolumn{1}{l|}{0.8972} & 0.9295 \\ \cline{2-8} 
                                       & Deceptive Auxiliary   Annotations                    & \multicolumn{1}{l|}{0.8609} & \multicolumn{1}{l|}{0.8449} & 0.8770 & \multicolumn{1}{l|}{0.8950} & \multicolumn{1}{l|}{0.8730} & 0.9169 \\ \cline{2-8} 
                                       & Modifying Legend                                     & \multicolumn{1}{l|}{0.8760} & \multicolumn{1}{l|}{0.8557} & 0.8962 & \multicolumn{1}{l|}{0.9017} & \multicolumn{1}{l|}{0.8808} & 0.9225 \\ \cline{2-8} 
                                       & Hiding Labels                                        & \multicolumn{1}{l|}{0.9077} & \multicolumn{1}{l|}{0.8770} & 0.9384 & \multicolumn{1}{l|}{0.9320} & \multicolumn{1}{l|}{0.9132} & 0.9507 \\ \cline{2-8} 
                                       & Adding or Removing Logos                             & \multicolumn{1}{l|}{0.8732} & \multicolumn{1}{l|}{0.8495} & 0.8970 & \multicolumn{1}{l|}{0.9145} & \multicolumn{1}{l|}{0.8958} & 0.9333 \\ \cline{2-8} 
                                       & Data-Visual Disproportion                            & \multicolumn{1}{l|}{0.8815} & \multicolumn{1}{l|}{0.8608} & 0.9022 & \multicolumn{1}{l|}{0.9040} & \multicolumn{1}{l|}{0.8816} & 0.9264 \\ \cline{2-8} 
                                       & Modifying Colormap                                   & \multicolumn{1}{l|}{0.8443} & \multicolumn{1}{l|}{0.8206} & 0.8681 & \multicolumn{1}{l|}{0.8933} & \multicolumn{1}{l|}{0.8739} & 0.9126 \\ \cline{2-8} 
                                       & Mixture                                              & \multicolumn{1}{l|}{0.9063} & \multicolumn{1}{l|}{0.8922} & 0.9203 & \multicolumn{1}{l|}{0.9162} & \multicolumn{1}{l|}{0.8942} & 0.9381 \\ \hline
\multirow{10}{*}{Baseline}             & Modifying Data Point Values                          & \multicolumn{1}{l|}{0.8407} & \multicolumn{1}{l|}{0.8208} & 0.8605 & \multicolumn{1}{l|}{0.8983} & \multicolumn{1}{l|}{0.8743} & 0.9223 \\ \cline{2-8} 
                                       & Adding or Removing Data Points                       & \multicolumn{1}{l|}{0.8243} & \multicolumn{1}{l|}{0.7863} & 0.8622 & \multicolumn{1}{l|}{0.8546} & \multicolumn{1}{l|}{0.8277} & 0.8815 \\ \cline{2-8} 
                                       & Modifying Coordinate Values                          & \multicolumn{1}{l|}{0.8283} & \multicolumn{1}{l|}{0.7954} & 0.8611 & \multicolumn{1}{l|}{0.8670} & \multicolumn{1}{l|}{0.8409} & 0.8931 \\ \cline{2-8} 
                                       & Deceptive Auxiliary   Annotations                    & \multicolumn{1}{l|}{0.8600} & \multicolumn{1}{l|}{0.8438} & 0.8762 & \multicolumn{1}{l|}{0.8998} & \multicolumn{1}{l|}{0.8860} & 0.9136 \\ \cline{2-8} 
                                       & Modifying Legend                                     & \multicolumn{1}{l|}{0.8592} & \multicolumn{1}{l|}{0.8398} & 0.8787 & \multicolumn{1}{l|}{0.8802} & \multicolumn{1}{l|}{0.8565} & 0.9040 \\ \cline{2-8} 
                                       & Hiding Labels                                        & \multicolumn{1}{l|}{0.8353} & \multicolumn{1}{l|}{0.8091} & 0.8614 & \multicolumn{1}{l|}{0.8914} & \multicolumn{1}{l|}{0.8638} & 0.9191 \\ \cline{2-8} 
                                       & Adding or Removing Logos                             & \multicolumn{1}{l|}{0.7897} & \multicolumn{1}{l|}{0.7555} & 0.8240 & \multicolumn{1}{l|}{0.8692} & \multicolumn{1}{l|}{0.8422} & 0.8961 \\ \cline{2-8} 
                                       & Data-Visual Disproportion                            & \multicolumn{1}{l|}{0.8643} & \multicolumn{1}{l|}{0.8423} & 0.8863 & \multicolumn{1}{l|}{0.8886} & \multicolumn{1}{l|}{0.8627} & 0.9145 \\ \cline{2-8} 
                                       & Modifying Colormap                                   & \multicolumn{1}{l|}{0.8375} & \multicolumn{1}{l|}{0.8066} & 0.8684 & \multicolumn{1}{l|}{0.8797} & \multicolumn{1}{l|}{0.8655} & 0.8939 \\ \cline{2-8} 
                                       & Mixture                                              & \multicolumn{1}{l|}{0.8189} & \multicolumn{1}{l|}{0.8031} & 0.8346 & \multicolumn{1}{l|}{0.8820} & \multicolumn{1}{l|}{0.8694} & 0.8945 \\ \hline
\end{tabular}

\end{table}

\vfill
\newpage

\section{Visual Comparison}
\begin{figure}[!h]
\centering
\setlength{\fboxrule}{0.4pt}
\setlength{\fboxsep}{0cm}
\includegraphics[width=.9\linewidth]{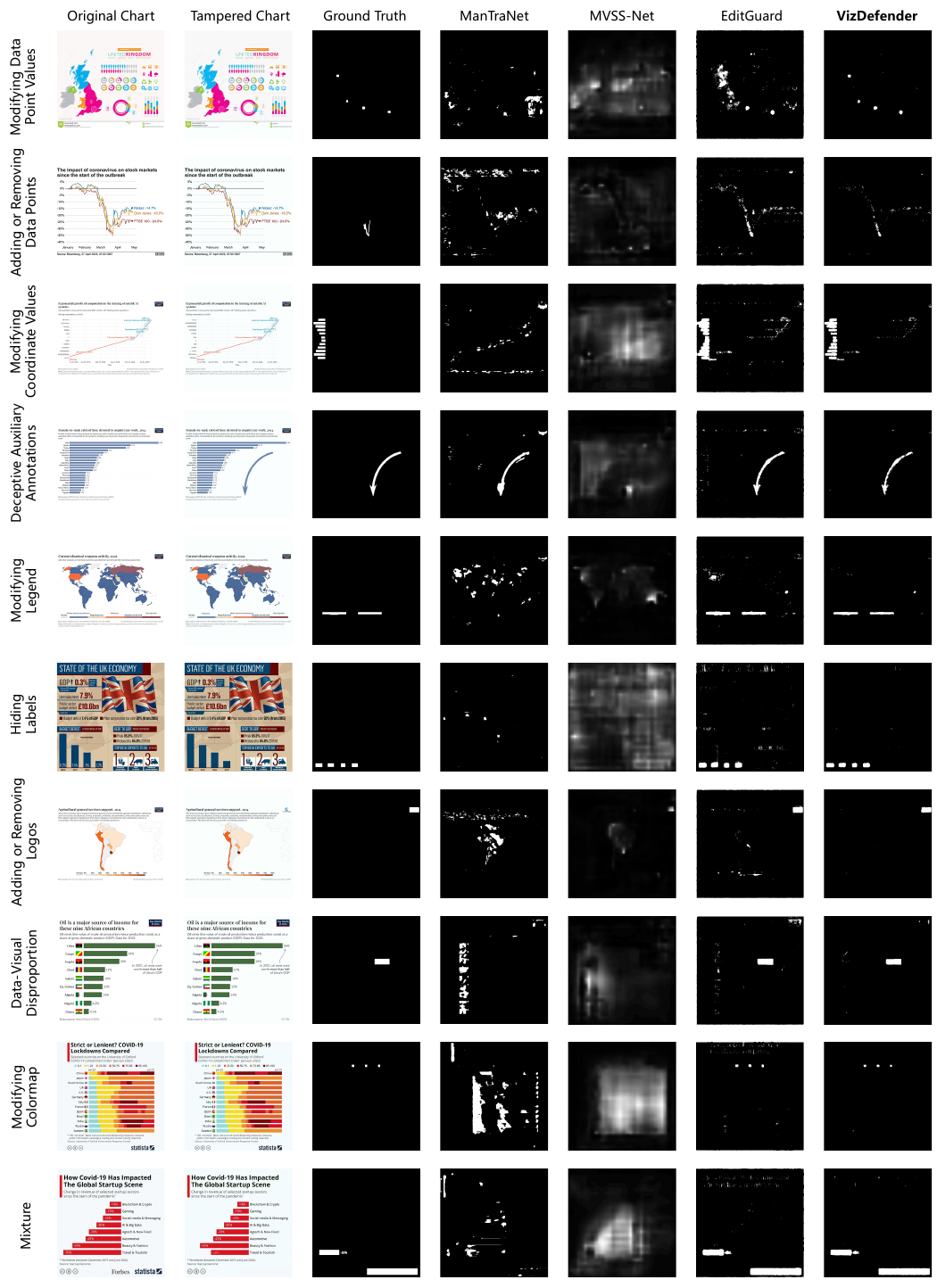}\ 
\caption{\label{fig:vis_comp_mcd} Visual comparison of tampering detection on the MCD. 
}
\end{figure}

\vfill
\newpage
\begin{figure}[!h]
\centering
\setlength{\fboxrule}{0.4pt}
\setlength{\fboxsep}{0cm}
\includegraphics[width=0.9\linewidth]{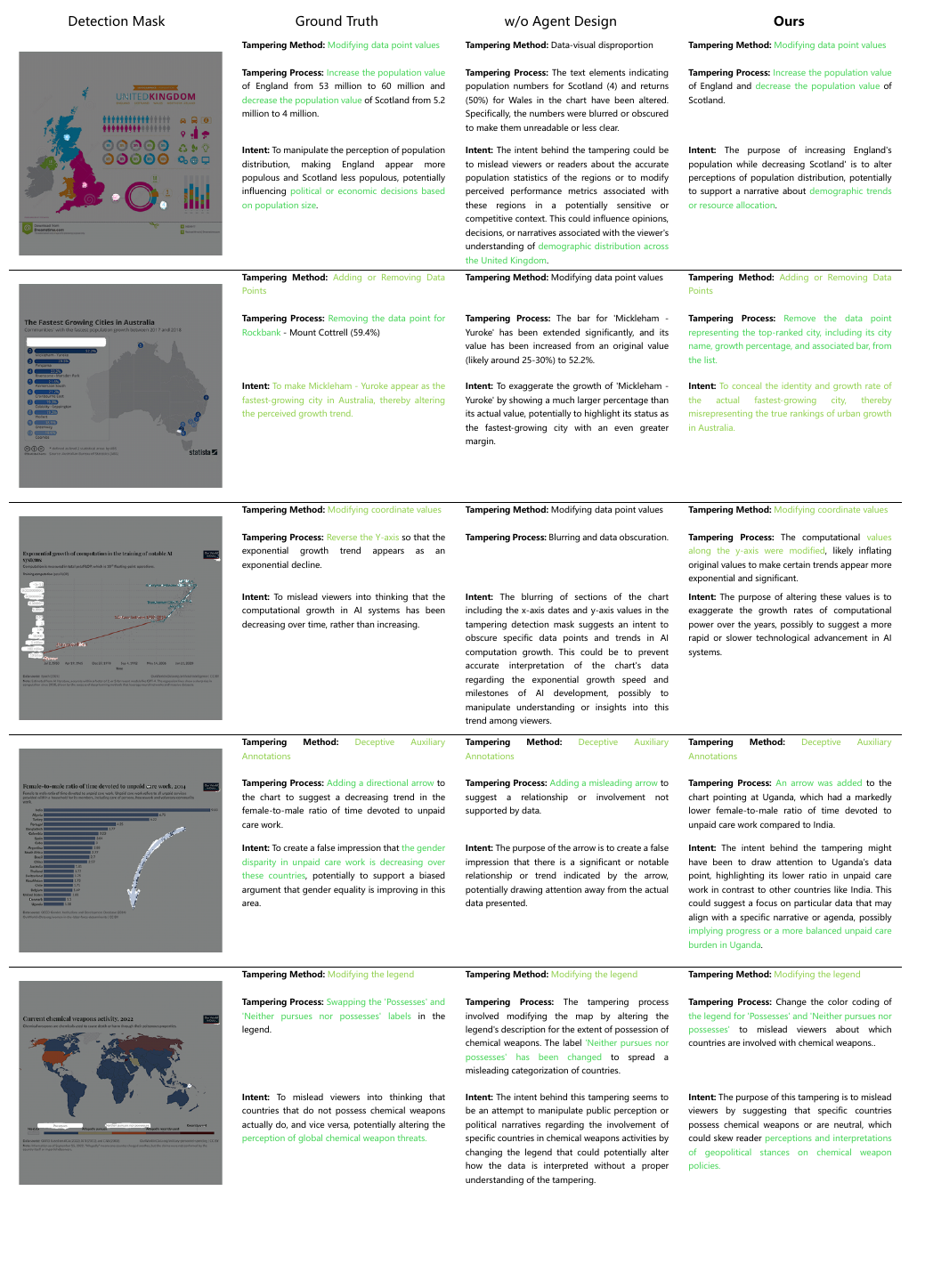}\ 
\caption{\label{fig:vis_comp_intent_1} Visual comparison of intent analysis on the AGD (the first five subsets).
}
\end{figure}

\vfill
\newpage\begin{figure}[!h]
\centering
\setlength{\fboxrule}{0.4pt}
\setlength{\fboxsep}{0cm}
\includegraphics[width=0.9\linewidth]{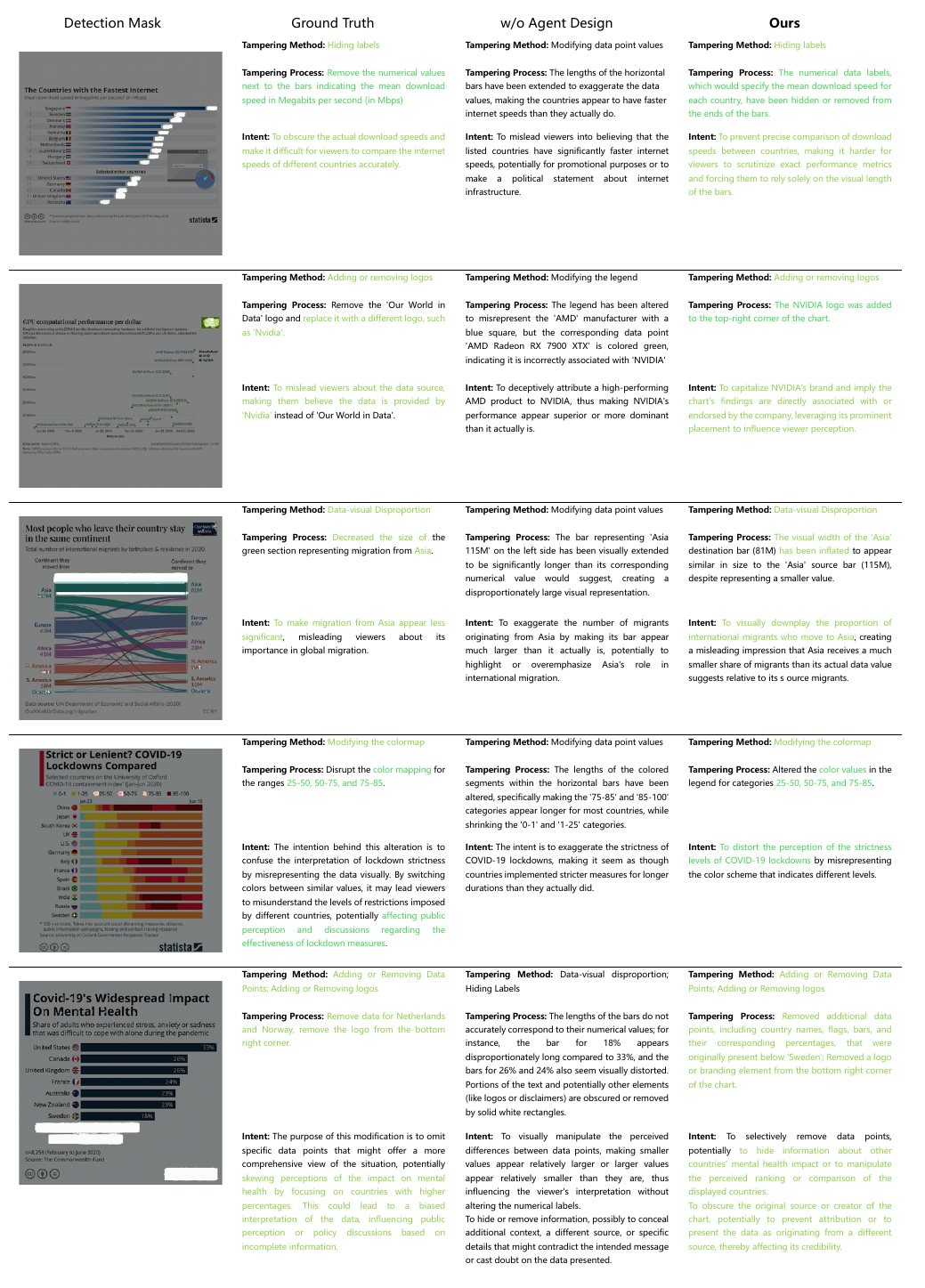}\ 
\caption{\label{fig:vis_comp_intent_2} Visual comparison of intent analysis on the AGD (the second five subsets).
}
\end{figure}

\vfill
\newpage

\begin{figure}[!h]
\centering
\setlength{\fboxrule}{0.4pt}
\setlength{\fboxsep}{0cm}
\includegraphics[width=0.99\linewidth]{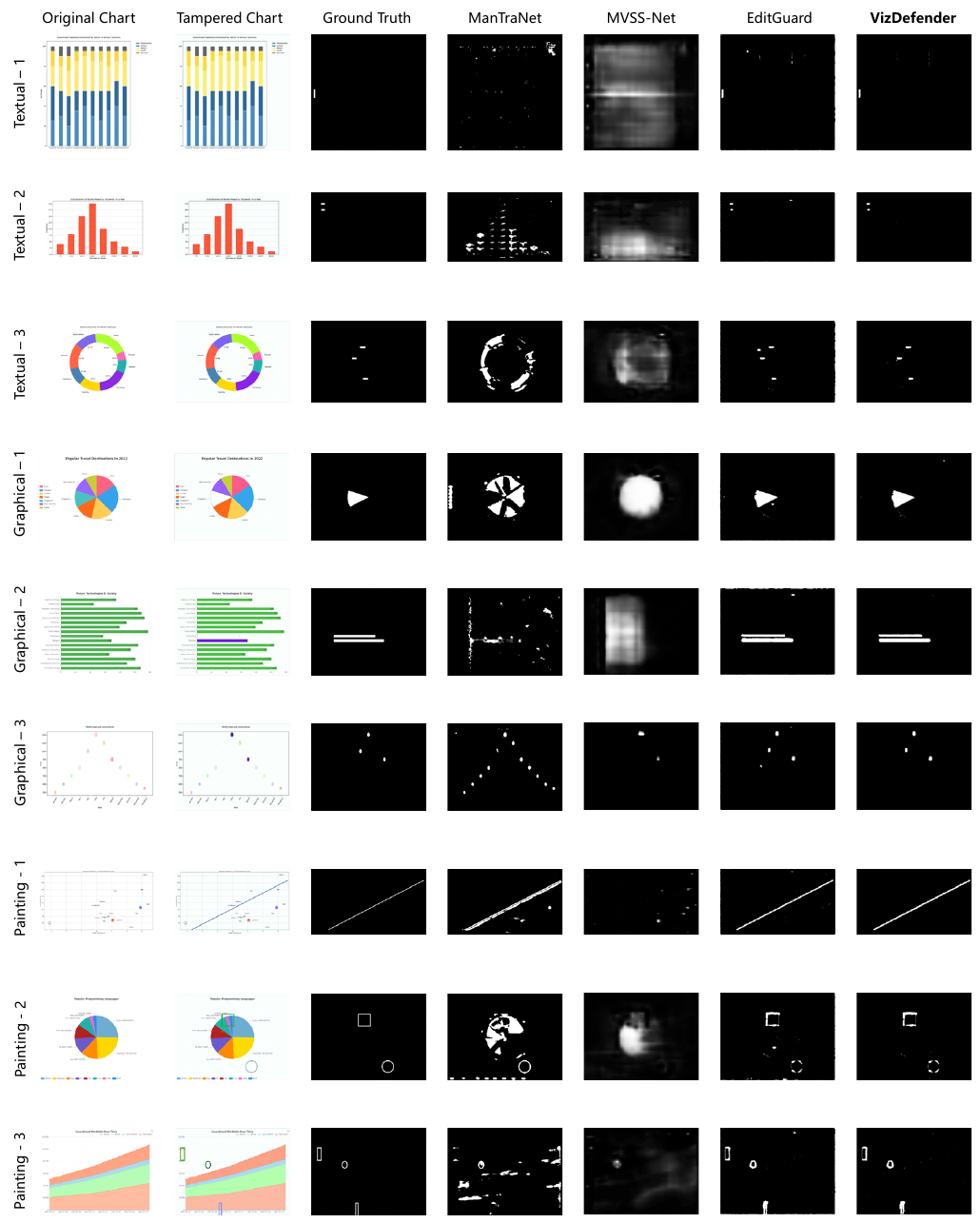}\ 
\caption{\label{fig:vis_comp_agd} Visual comparison of tampering detection on the AGD.
}
\end{figure}

\vfill
\newpage

\section{Failure Analysis}
\subsection{Tampering Detection}
The current tampering detection method in the VizDefender system still faces certain failure cases. By analyzing two primary errors, we find that these issues collectively highlight the model's insufficient sensitivity to fine details, particularly in thin-line structures. Most of these noise artifacts can be effectively addressed through our Mask Refinement Agent.

\begin{figure}[!h]
\centering
\setlength{\fboxrule}{0.4pt}
\setlength{\fboxsep}{0cm}
\includegraphics[width=0.8\linewidth]{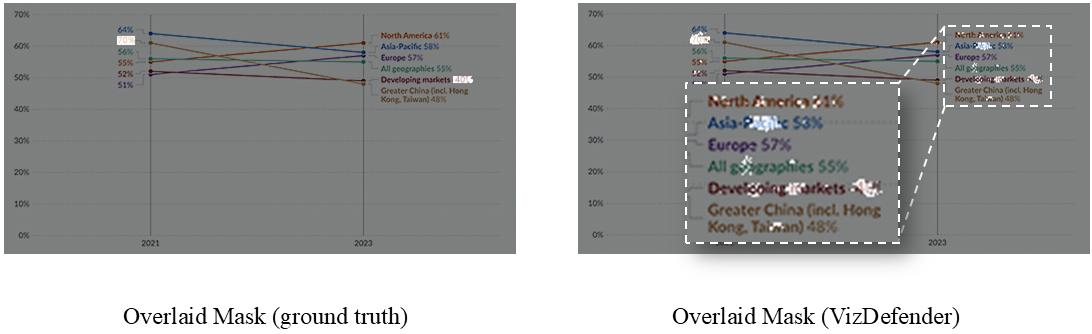}\ 
\caption{\label{fig:failure_analysis_1} Error map of Noise in Text Region Detection. The rectangular boxes in the figure represent the locations where errors occur.
}
\end{figure}

\textbf{Noise in Text Region Detection.}
As shown in Fig.~\ref{fig:failure_analysis_1}, in the process of detecting tampering in text regions, the model often generates extraneous noise responses at the edges of characters. This results in false-positive outputs that resemble burr-like anomalies, which degrade the accuracy of detecting true tampered regions. These artifacts can lead to misjudgments, especially in cases where the tampered regions are subtle or where the noise is incorrectly identified as tampering.

The error may stem from the fact that visualization images, in contrast to natural images, often include large areas of blank space with high brightness. During the watermarking process, we used DWT, which primarily targets embedding watermarks in high-frequency regions. For visualizations with large blank areas, the high-frequency components are concentrated in the text regions. Since the amount of watermarked information is fixed, the watermark is prone to damage during file transmission, leading to noise appearing more frequently in text regions.

\begin{figure}[!h]
\centering
\setlength{\fboxrule}{0.4pt}
\setlength{\fboxsep}{0cm}
\includegraphics[width=0.8\linewidth]{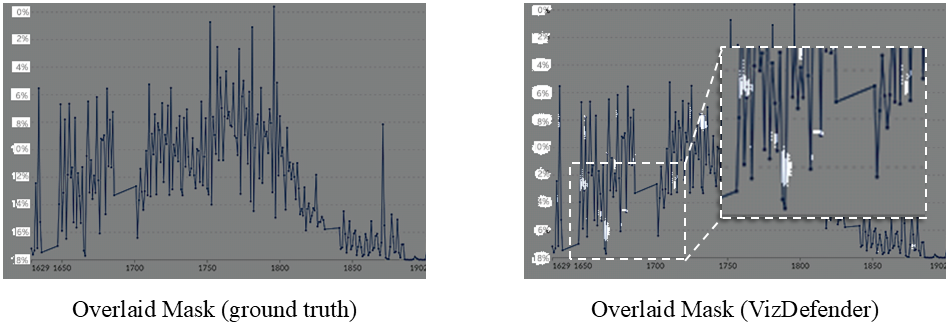}\ 
\caption{\label{fig:failure_analysis_2} Error map of False Detection of Thin-Line Structures. The rectangular boxes in the figure represent the locations where errors occur.
}
\end{figure}

\textbf{False Detection of Thin-Line Structures.}
As shown in Fig.~\ref{fig:failure_analysis_2}, for images containing thin-line structures, such as lines in line charts, the model tends to produce noise along the lines themselves. These noise artifacts are frequently mistaken for genuine tampering traces, such as erasures or additions of lines. As a result, the system's ability to discern between natural features and tampered areas is compromised, leading to incorrect detection of tampering in the absence of such changes.

The error may occur due to thin lines being vulnerable to information loss during the convolution process. During the deconvolution process, these fine details are often further degraded, leading to the loss of critical information and resulting in noise artifacts along the lines.

\vfill
\newpage

\subsection{Tampering Method and Intent Analysis Inference}
\begin{figure}[!h]
\centering
\setlength{\fboxrule}{0.4pt}
\setlength{\fboxsep}{0cm}
\includegraphics[width=\linewidth]{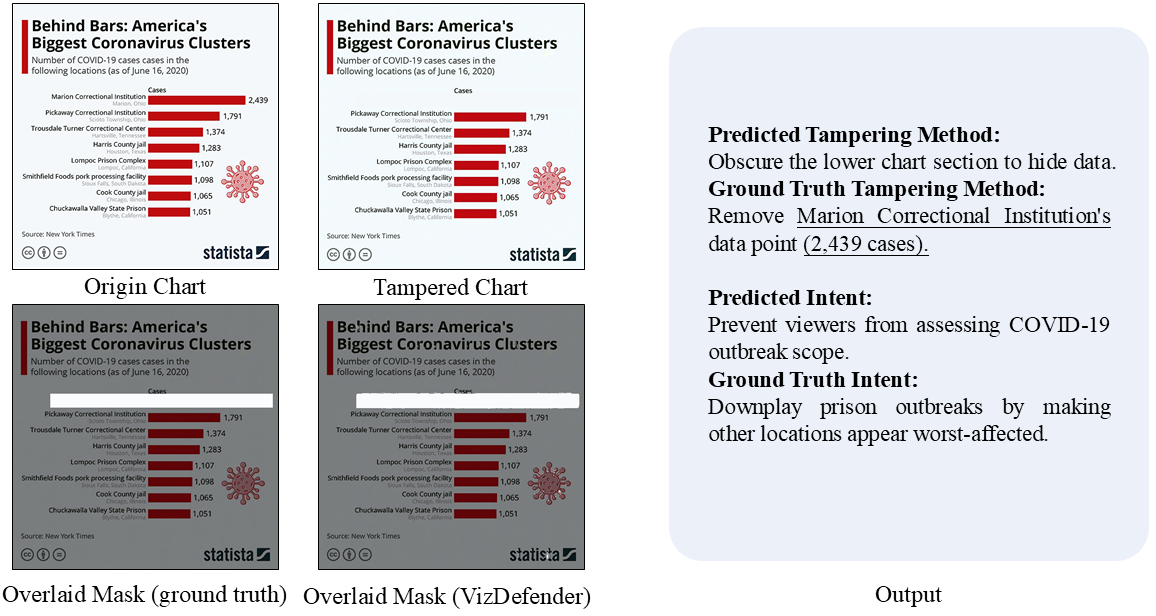}\ 
\caption{\label{fig:failure_analysis_6}  Illustration of tampering intent inference when prior knowledge is absent. The horizontal line in the tampered visualization represents the lack of prior knowledge about the removed data point, leading to the model misinterpreting the tampering method and intent.
}
\end{figure}

The current tampering method and intent analysis inference in the VizDefender system faces challenges when tampering involves domain-specific references or context. The agent has difficulty leveraging prior knowledge to infer tampering method or intent because of the lack of the original visualization. As shown in Fig.~\ref{fig:failure_analysis_6}, when tampering involves specific locations, such as ``\textit{Marion Correctional Institution with 2,439 cases}'', the agent lacks the information of this target. This absence of prior knowledge results in the model incorrectly predicting the tampering intent as simply obscuring the lower section of the visualization to hide data, instead of recognizing the true manipulation, which involves removing the data for Marion Correctional Institution. Consequently, the model misinterprets both the tampering method and intent, failing to accurately identify the underlying manipulation.

This limitation is not only an issue for VizDefender but also for human judgment. In cases where individuals lack the necessary domain knowledge or context, similar misinterpretations can occur. Therefore, a potential solution for future work is to leverage the prior visualization information into the images.

\end{document}